\documentclass{nm}
\renewcommand\thisnumber{x}
\renewcommand\thisyear {2026}
\renewcommand\thismonth{x}
\renewcommand\thisvolume{xx}
\renewcommand\doi{10.4208/nmtma.  }
\usepackage{charter}
\usepackage{booktabs}
\usepackage{multirow}
\usepackage[table]{xcolor}
\usepackage{float}
\usepackage{url}
\usepackage{booktabs}
\usepackage{longtable}
\usepackage{array}

\usepackage{caption}
\usepackage{subcaption}
\usepackage[hidelinks]{hyperref}

\begin{document}

\markboth{    }{      }
\title{Multi-scale Multiple Scattering Neural Operator for Inverse Medium Problems}



\author[R. Ming et al.]{
Rong Ming\affil{1},
Kuilin Qin\affil{1},
Jiwei Jia\affil{2},
Yu Wang\affil{3} 
and Yuping Duan\affil{1}\comma\corrauth
}

\address{
\affilnum{1}
School of Mathematical Sciences, Beijing Normal University,
Beijing 100875, China.
\\
\affilnum{2}
Department of Mathematics, Jilin University,
Changchun 130015, China.
\\
\affilnum{3}
Key Laboratory of Digital Technology in Medical Diagnostics of Zhejiang,
Hangzhou 310030, China.
}

\emails{
\raggedright
\texttt{doveduan@gmail.com} (Y. Duan),
\texttt{202431130070@mail.bnu.edu.cn} (R. Ming),
\texttt{202321130113@mail.bnu.edu.cn} (K. Qin),
\texttt{jiajiwei@jlu.edu.cn} (J. Jia),
\texttt{flimanadam@gmail.com} (Yu Wang),
}

\begin{abstract}
Accurate wavefield prediction and fine-scale medium reconstruction remain challenging for neural operators, in the presence of multiple scattering and rapidly oscillating wavefields. We propose a latent-space Multiple Scattering Neural Operator (MSNO) that incorporates the recursive structure of the Born iteration for the Helmholtz equation into a learnable forward model. MSNO approximates the mapping from the medium contrast and incident field to the total wavefield through spectral interactions and recursive latent updates with shared parameters. An auxiliary autoencoder loss encourages encoder--decoder reconstruction consistency for the incident field. We develop a multiscale extension, MS-MSNO, that combines scale-conditioned branches with spectrally adaptive fusion. The fusion weights are computed for each input from the radial spectral responses of the branch predictions, enabling sample-dependent aggregation. The differentiable forward surrogate supports gradient-based medium reconstruction without separate forward and adjoint PDE solves during inversion. Experiments on synthetic electromagnetic and acoustic breast-phantom datasets show improved forward prediction and inverse reconstruction relative to the FNO and BFNO baselines. Spectral analyses and ablation studies support improved approximation of upper-frequency wavefield components and recovery of fine-scale structures. In the breast-phantom experiment, MS-MSNO achieves these accuracy gains with a parameter count and training time per epoch comparable to those of single-scale FNO.
\end{abstract}



\maketitle

\section{Introduction}
Wave scattering in inhomogeneous media provides the physical and mathematical foundation for inverse medium problems, which aim to recover internal material properties from measured wavefields under prescribed excitations. Depending on the physical model, the target parameters include sound speed and density in acoustic scattering, as well as electric permittivity and conductivity in electromagnetic scattering. Inverse medium problems play a central role in a wide range of imaging and scientific applications. In medical imaging, ultrasound tomography can be formulated as an inverse scattering problem, in which tissue acoustic properties are reconstructed from wave measurements acquired around the object, supporting non-invasive tissue characterization and quantitative imaging. In geophysical exploration \cite{tarantola1984inversion}, seismic wave scattering is used to infer subsurface structures and material properties, forming the theoretical basis of seismic imaging and full-waveform inversion for resource exploration. Furthermore, in electromagnetic and optical imaging \cite{colton1998inverse}, inverse scattering theory underpins applications such as radar imaging and target characterization, where information about objects and material properties is inferred from scattered electromagnetic fields.

The scalar Helmholtz equation provides a standard model for time-harmonic wave propagation under appropriate physical assumptions. In two-dimensional lossless media, it takes the form
\begin{equation}
    \Delta u(x)+k^2\bigl(1+q(x)\bigr)u(x)=s(x),
    \qquad x\in\mathbb{R}^2,
    \label{eq:helmholtz_intro}
\end{equation}
where $u$ is the complex-valued total field, $s$ is a prescribed source term, $k>0$ is the background wavenumber, and $q$ is a real-valued, dimensionless medium contrast relative to a homogeneous background. The physical interpretation of $q$ depends on the application. For electromagnetic scattering with constant magnetic permeability, \(q\) represents the contrast associated with the dielectric properties of the medium. For constant-density acoustic scattering, \(q\) characterizes the contrast induced by spatial variations of the sound-speed distribution. The total field is decomposed as $u=u_{\mathrm{in}}+u_{\mathrm{sca}}$, where $u_{\mathrm{in}}$ is the incident field and $u_{\mathrm{sca}}$ is the scattered field induced by the medium contrast. For a fixed contrast, the governing equation is linear in the wavefield, whereas the total field generally depends nonlinearly on the contrast.

Solving forward Helmholtz scattering problems presents significant numerical challenges because of the oscillatory nature of high-frequency wavefields and the unbounded propagation domain. A common strategy is to truncate the physical domain using absorbing layers or artificial boundary conditions, thereby reformulating the problem on a bounded computational domain. Widely adopted techniques include perfectly matched layers (PMLs) \cite{du2022nonlocal,8} and absorbing boundary conditions (ABCs) \cite{11,12}, which approximate outgoing wave propagation while limiting spurious reflections. After domain truncation, classical schemes such as the finite difference method (FDM) \cite{FDM1,FDM2} and the finite element method (FEM) \cite{FEM,FEM1} are used to discretize the resulting boundary value problem. High-order compact finite-difference schemes have also been developed to improve spatial accuracy for Poisson and Helmholtz equations on structured grids \cite{pan2023sixth}. Finite element methods, in particular, offer flexibility in handling complex geometries and heterogeneous media. At high frequencies, resolving wave oscillations and controlling numerical dispersion can require fine spatial discretization, while the indefinite nature of Helmholtz systems poses additional challenges for iterative solvers and preconditioning \cite{ernst2011difficult}. The resulting large-scale systems can incur substantial computational and memory costs, particularly when repeated forward solves are required during inversion. An alternative is the Lippmann--Schwinger formulation \cite{L-S}, which expresses the scattering problem as a volume integral equation over the support of the medium contrast. By incorporating the outgoing background Green's function, this formulation accounts for propagation in the homogeneous exterior without an artificial outer boundary. Although discretization generally produces dense interactions, the convolution structure of the background Green's function permits fast evaluation on suitable grids. Such approaches have been extensively studied in acoustic and electromagnetic scattering. Furthermore, hybrid finite element--boundary integral methods combine a local discretization of the heterogeneous region with a nonlocal representation of the exterior field \cite{jin1991finite}. Related coupled formulations have also been analyzed in specific direct and inverse time-harmonic scattering settings, including their well-posedness and local stability \cite{bao2011cavity}.

With its capacity to approximate complex nonlinear mappings, deep learning has been increasingly applied to high-dimensional modeling and the numerical solution of partial differential equations. Alongside empirical developments, theoretical studies have analyzed approximation and statistical errors for specific classes of neural-network-based PDE solvers \cite{chen2024deepritz}. In inverse problems, end-to-end data-driven methods based on convolutional neural networks (CNNs) \cite{lecun2002gradient} learn direct mappings from measurement data to medium parameters. For example, InversionNet \cite{wu2019inversionnet} employs a deep convolutional architecture to map seismic observations to subsurface velocity models. Encoder--decoder architectures such as U-Net \cite{ronneberger2015u} provide a general framework for image-to-image prediction, while generative models have also been used to represent geological priors in seismic inversion \cite{mosser2020stochastic}. Once trained, direct reconstruction networks can produce estimates efficiently. However, purely data-driven implementations do not necessarily enforce consistency with the governing equations, and their accuracy may deteriorate when the medium distribution or acquisition configuration differs from the training setting.

Operator learning methods seek to approximate mappings between function spaces, providing a framework for learning solution operators of parameterized PDEs. DeepONet \cite{lu2019deeponet} combines a branch network that encodes sampled input functions with a trunk network that encodes evaluation coordinates. The Fourier Neural Operator (FNO) \cite{FNO} represents nonlocal interactions through learned Fourier multipliers. Related architectures include the Adaptive Fourier Neural Operator (AFNO) \cite{guibas2021adaptive}, which introduces structured mixing in the Fourier domain, and the Wavelet Neural Operator (WNO) \cite{tripura2023wavelet}, which uses wavelet representations to capture spatially localized features at multiple scales. For Helmholtz-type problems, the Born Fourier Neural Operator (BFNO) \cite{BFNO} and related methods \cite{nsno,neumann} incorporate recursive structures inspired by Born or Neumann series to model successive wave--medium interactions through learned operator updates. These architectures establish an explicit connection to classical scattering iterations, although this structural connection alone does not guarantee physical consistency or convergence of the learned updates.

\begin{figure}[t]    
  \centering            
  \subfloat[]   
  {
      \label{fig1:1}\includegraphics[width=0.44\textwidth]{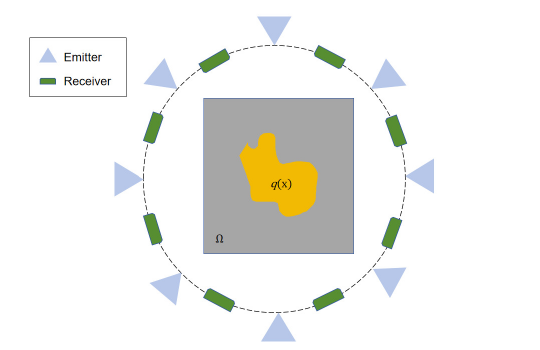}
  }
  \subfloat[]
  {
      \label{fig1:2}\includegraphics[width=0.5\textwidth]{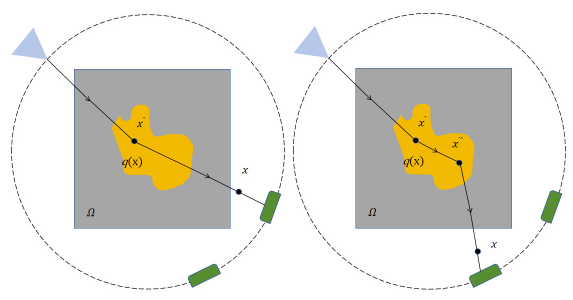}
  }
  \caption{(a) Schematic diagram of the inverse medium problem; (b) Illustration of single scattering and double scattering.}    
  \label{fig1}            
\end{figure}

As illustrated in Figure~\ref{fig1}(a), the inverse medium problem seeks to recover the contrast $q$ from total-field measurements $\{d_j\}_{j=1}^{M}$ corresponding to prescribed incident fields $\{u_{\mathrm{in},j}\}_{j=1}^{M}$. At a given frequency, let $\mathcal{T}(u_{\mathrm{in},j};q)$ denote the total field obtained by solving the forward scattering problem, and let $\mathcal{R}$ restrict this field to the receiver locations. The reconstruction can be formulated as
\begin{equation}
\min_{q\in\mathcal{Q}_{\mathrm{ad}}}\mathcal{J}(q),
\qquad
\mathcal{J}(q)=\frac{1}{2}\sum_{j=1}^{M}
\left\|
\mathcal{R}\mathcal{T}(u_{\mathrm{in},j};q)-d_j
\right\|_2^2,
\label{eq:inverse_problem}
\end{equation}
where $\mathcal{Q}_{\mathrm{ad}}$ is the admissible set of contrasts and $\|\cdot\|_2$ denotes the Euclidean norm over the receiver data. Gradient-based reconstruction requires repeated forward and gradient evaluations, making numerical PDE solves a major computational cost. Differentiable neural surrogates approximate the forward operator and enable gradient computation through automatic differentiation \cite{paszke2017automatic}, avoiding separate forward and adjoint PDE solves during inversion. The resulting gradients correspond to the surrogate-based objective, so reconstruction quality depends in part on the accuracy of the learned approximation.

Inspired by the recursive structure of the Born iteration, we propose a latent-space Multiple Scattering Neural Operator (MSNO) for frequency-domain wavefield prediction in multiple-scattering regimes. Our MSNO approximates the forward map from the medium contrast and incident field to the total wavefield, including its nonlinear dependence on the medium contrast. The learned operator serves as a differentiable surrogate within a data-consistency-based inversion framework for reconstructing the medium from measured wavefields. To sum up, the main contributions are summarized as follows:
\begin{itemize}
    \item We propose MSNO with recursive latent updates conditioned on the medium representation and parameters shared across iterations. An auxiliary autoencoder loss promotes encoder--decoder reconstruction consistency for the incident field.

    \item We develop MS-MSNO with parallel scale-conditioned branches and spectrally adaptive fusion. Sample-dependent fusion weights are computed from the radial spectral responses of the current branch predictions in every forward pass during training and inference, including inversion. The model improves high-frequency wavefield approximation and fine-structure recovery in the reported experiments.

    \item Experiments on synthetic electromagnetic and acoustic breast-phantom datasets show improved forward prediction and inverse reconstruction relative to the tested baselines. In the reported breast-phantom configuration, MS-MSNO achieves these improvements with a parameter count and training time per epoch comparable to those of single-scale FNO.
\end{itemize}

The remainder of this paper is organized as follows. Section~\ref{sec2} presents the Helmholtz scattering formulation and the associated Born iteration. Section~\ref{sec3} develops the latent-space multiple scattering neural operator. Section~\ref{sec:ms_msno} introduces its multiscale extension with parallel branches and spectrally adaptive fusion. Section~\ref{sec5} presents the numerical experiments, and Section~\ref{sec6} concludes the paper.  For clarity and ease of reference, the key mathematical symbols
used throughout this paper are summarized in Table~\ref{tab:symbols}.

\begin{table}[h]
\centering
\small
\renewcommand{\arraystretch}{1.12}
\caption{Key notation for the MSNO and MS-MSNO architectures.}
\label{tab:symbols}
\begin{tabular}{@{}
    >{\raggedright\arraybackslash}p{0.30\textwidth}
    >{\raggedright\arraybackslash}p{0.65\textwidth}
    @{}}
\toprule
\textbf{Symbol} & \textbf{Description} \\
\midrule

$q$
& Medium contrast. \\

$u,u_{\mathrm{in}},u_{\mathrm{sca}}$
& Total, incident, and scattered fields. \\

$u_\theta$
& Predicted total field. \\

$\mathcal{S}_q,\mathcal{S}_\theta$
& Physical and learned latent scattering operators. \\

$P_{u_\theta},P_{q_\theta},Q_\theta$
& Incident-field encoder, contrast encoder, and field decoder. \\

$v_0,v_q,v_n$
& Incident-field embedding, medium embedding, and latent wavefield
state after $n$ updates. \\

$R_\Phi,\mathcal{M}$
& Learned spectral multiplier and pointwise transition block. \\

$\alpha_b$
& Input scaling factor of branch $b$. \\

$u_\theta^{(b)},\gamma_b$
& Total-field prediction and fusion coefficient of branch $b$. \\

\bottomrule
\end{tabular}
\end{table}

\section{Preliminaries} \label{sec2}

We formulate the scattering problem and its integral representation. Let $D\subset\mathbb{R}^2$ be a bounded domain containing the compact support of a real-valued contrast $q\in L^\infty(\mathbb{R}^2;\mathbb{R})$.
Assume that $1+q\geq a_0>0$ almost everywhere for some constant $a_0$. Throughout this section, $k>0$ is a fixed background wavenumber. The wavefields are complex-valued, and $L^2(D)$ denotes $L^2(D;\mathbb{C})$.
Let $u_{\mathrm{in}}$ be a prescribed incident field satisfying $(\Delta+k^2)u_{\mathrm{in}}=s$ in the homogeneous background. We assume that the restriction of $u_{\mathrm{in}}$ to $D$ belongs to $L^2(D)$ and that $u_{\mathrm{in}}$ is smooth in a neighborhood of the support of $q$.
Substituting $u=u_{\mathrm{in}}+u_{\mathrm{sca}}$ into Eq.~\eqref{eq:helmholtz_intro} and imposing the outgoing Sommerfeld radiation condition on $u_{\mathrm{sca}}$ gives the scattering problem
\begin{equation}
\begin{cases}
\Delta u_{\mathrm{sca}}(x)
+k^2\bigl(1+q(x)\bigr)u_{\mathrm{sca}}(x)
=-k^2q(x)u_{\mathrm{in}}(x),
& x\in\mathbb{R}^2, \\[4pt]
\displaystyle
\lim_{r\to\infty}\sqrt{r}
\left(
\frac{\partial u_{\mathrm{sca}}}{\partial r}
-\mathrm{i}k u_{\mathrm{sca}}
\right)=0,
& r=|x|.
\end{cases}
\label{eq:scattering_system}
\end{equation}
The differential equation is understood in the distributional sense, and the radiation limit holds uniformly with respect to the observation direction $x/|x|$. The Sommerfeld radiation condition selects outgoing scattered waves. Under the stated assumptions, the problem admits a unique radiating solution
$u_{\mathrm{sca}}\in H^1_{\mathrm{loc}}(\mathbb{R}^2;\mathbb{C})$
\cite{colton1998inverse}. Existence follows from Fredholm theory for the associated integral equation together with uniqueness of the homogeneous scattering problem.

Using the outgoing fundamental solution of the background Helmholtz operator, we obtain the Lippmann--Schwinger (LS) integral equation for the total field:
\begin{equation}
u(x)
=
u_{\mathrm{in}}(x)
+k^2\int_D \Phi(x,y)q(y)u(y)\,\mathrm{d}y,
\qquad x\in D.
\label{eq:LS}
\end{equation}
In two dimensions, the outgoing free-space Green's function is
\[
\Phi(x,y)
=
\frac{\mathrm{i}}{4}
H_0^{(1)}\bigl(k|x-y|\bigr),
\qquad x\neq y,
\]
where $H_0^{(1)}$ is the Hankel function of the first kind of order zero. With this sign convention,
\[
(\Delta_x+k^2)\Phi(x,y)=-\delta_y(x)
\]
holds in the distributional sense, where $\delta_y$ is the Dirac distribution centered at $y$.
The dependence of $\Phi$ on the fixed wavenumber $k$ is suppressed in the notation.
Equation~\eqref{eq:LS} is understood as an identity in $L^2(D)$. Once the total field in $D$ is determined, the volume-potential term in Eq.~\eqref{eq:LS} determines the scattered field throughout $\mathbb{R}^2$. This representation couples the medium contrast with the total field inside $D$ and provides the structural basis for the proposed neural operator.

To make the medium dependence explicit, we define the scattering integral operator
$\mathcal{S}_q:L^2(D)\to L^2(D)$ by
\begin{equation}
(\mathcal{S}_q v)(x)
:=
k^2\int_D \Phi(x,y)q(y)v(y)\,\mathrm{d}y,
\qquad x\in D.
\label{eq:S}
\end{equation}
For fixed $q$, the operator $\mathcal{S}_q$ is linear and compact on $L^2(D)$. Indeed, the logarithmic singularity of $\Phi$ in two dimensions, together with the boundedness of $D$ and $q$, ensures that its integral kernel is square-integrable on $D\times D$.
The dependence of $\mathcal{S}_q$ on $k$ is also suppressed. In the following operator identities, $u$ and $u_{\mathrm{in}}$ denote their restrictions to $D$.
Then, Eq.~\eqref{eq:LS} becomes
\[
(I-\mathcal{S}_q)u=u_{\mathrm{in}},
\]
where $I$ is the identity operator on $L^2(D)$. Under the stated scattering assumptions, uniqueness and the Fredholm alternative imply that $I-\mathcal{S}_q$ is invertible. The associated Born iteration is
\begin{equation}
u_{n+1}
=
u_{\mathrm{in}}+\mathcal{S}_q u_n,
\qquad n=0,1,\ldots,
\qquad u_0=u_{\mathrm{in}}.
\label{eq:born_iter_formal}
\end{equation}
After $N$ updates, the iterate is
\[
u_N=\sum_{n=0}^{N}\mathcal{S}_q^n u_{\mathrm{in}},
\]
where $\mathcal{S}_q^0=I$. If the spectral radius satisfies
$\rho(\mathcal{S}_q)<1$, the operator series
$\sum_{n=0}^{\infty}\mathcal{S}_q^n$
converges in operator norm to $(I-\mathcal{S}_q)^{-1}$. Consequently, $u_N$ converges in $L^2(D)$ to the total field $u$ as 
\begin{equation}
u
=
(I-\mathcal{S}_q)^{-1}u_{\mathrm{in}}
=
u_{\mathrm{in}}
+\sum_{n=1}^{N}\mathcal{S}_q^n u_{\mathrm{in}}, \quad \text{as }  N\rightarrow \infty.
\label{eq:Neumann}  
\end{equation}
A sufficient condition for convergence is
$\|\mathcal{S}_q\|_{L^2(D)\to L^2(D)}<1$.
For $n\geq1$, the term $\mathcal{S}_q^n u_{\mathrm{in}}$ represents an $n$-th order scattering contribution. The invertibility of $I-\mathcal{S}_q$ alone does not guarantee convergence of the Born series. Thus, well-posedness of the scattering problem does not imply convergence of the associated iteration.

\section{Multiple Scattering Neural Operator}\label{sec3}

The first-order Born approximation neglects higher-order scattering interactions, yielding a linear approximation of the dependence of the scattered field on the medium contrast for a prescribed incident field. Its accuracy can deteriorate when multiple scattering is significant. The classical Born iteration accounts for successive scattering interactions, but its convergence requires additional conditions, as discussed in Section~\ref{sec2}. Inspired by this recursive structure, we develop a Multiple Scattering Neural Operator (MSNO) that approximates the total-field solution map through medium-dependent updates in a latent feature space. The architecture combines spectral operations with a finite sequence of recursive updates whose parameters are shared across iterations. These learned updates follow the structure of the classical iteration without assuming that its convergence properties carry over to the neural architecture.

\subsection{Latent space modeling}
We perform spectral operations in a latent feature space to model successive wave--medium interactions. We use separate embedding operators to lift the incident field $u_{\mathrm{in}}(x)$ and the medium contrast $q(x)$ into corresponding feature representations. An auxiliary autoencoder loss encourages encoder--decoder reconstruction consistency for the incident field. The incident field is encoded as the initial state of the learned scattering recursion:
\[
v_0(x)
=
P_{u_{\theta}}\!\bigl(
x,\ \operatorname{Re}\{u_{\mathrm{in}}(x)\},\
\operatorname{Im}\{u_{\mathrm{in}}(x)\}
\bigr),
\]
where $P_{u_{\theta}}:\mathbb{R}^{4}\to\mathbb{R}^{d_v}$ is a multilayer perceptron (MLP) applied pointwise, and $d_v$ denotes the latent feature dimension. The four inputs comprise the two spatial coordinates of $x\in\mathbb{R}^2$ and the real and imaginary components of the incident field. The resulting embedding $v_0$ initializes the learned scattering recursion.

After $N$ recursive updates, the final latent state $v_N(x)$ is mapped to the field representation by a decoder
$Q_{\theta}:\mathbb{R}^{d_v}\to\mathbb{R}^{2}$.
Its two output channels represent the real and imaginary components of the predicted complex-valued total field $u_\theta$. Distinguishing this prediction from the physical solution $u$, we write
\[
\big[
\operatorname{Re}\{u_{\theta}(x)\},\
\operatorname{Im}\{u_{\theta}(x)\}
\big]
=
Q_{\theta}\bigl(v_N(x)\bigr),
\]
where the brackets denote concatenation along the channel dimension.

The auxiliary autoencoder loss encourages the same decoder $Q_{\theta}$ to reconstruct the incident field directly from its initial embedding $v_0$:
\[
Q_{\theta}\bigl(v_0(x)\bigr)
\approx
\big[
\operatorname{Re}\{u_{\mathrm{in}}(x)\},\
\operatorname{Im}\{u_{\mathrm{in}}(x)\}
\big].
\]
The reconstructed incident field uses the same channel representation as the predicted total field. This relation is encouraged by minimizing the reconstruction error over the training samples, rather than imposed as an exact pointwise identity. It neither expresses a physical conservation law nor guarantees reconstruction consistency at intermediate iterations. The recursive updates operate directly in latent space, without encoding and decoding between successive steps.

Similarly, the medium contrast is encoded through a pointwise transformation:
\[
v_q(x)=P_{q_{\theta}}\!\bigl(x,\ q(x)\bigr),
\]
where $P_{q_{\theta}}:\mathbb{R}^{3}\to\mathbb{R}^{d_v}$ is a pointwise MLP. Its three inputs comprise the two spatial coordinates of $x\in\mathbb{R}^2$ and the scalar contrast $q(x)$.
The medium embedding has the same feature dimension $d_v$ as the wavefield embedding, enabling channelwise interactions within each learned scattering update. For a given input sample, $v_q$ remains fixed throughout the recursion while the latent wavefield evolves. Thus, every update is conditioned on the same medium representation, reflecting the use of a prescribed contrast in the classical scattering iteration.

\subsection{Multiple scattering neural operator}

\begin{figure}[t]    
\centering    
\includegraphics[width=0.9\linewidth]{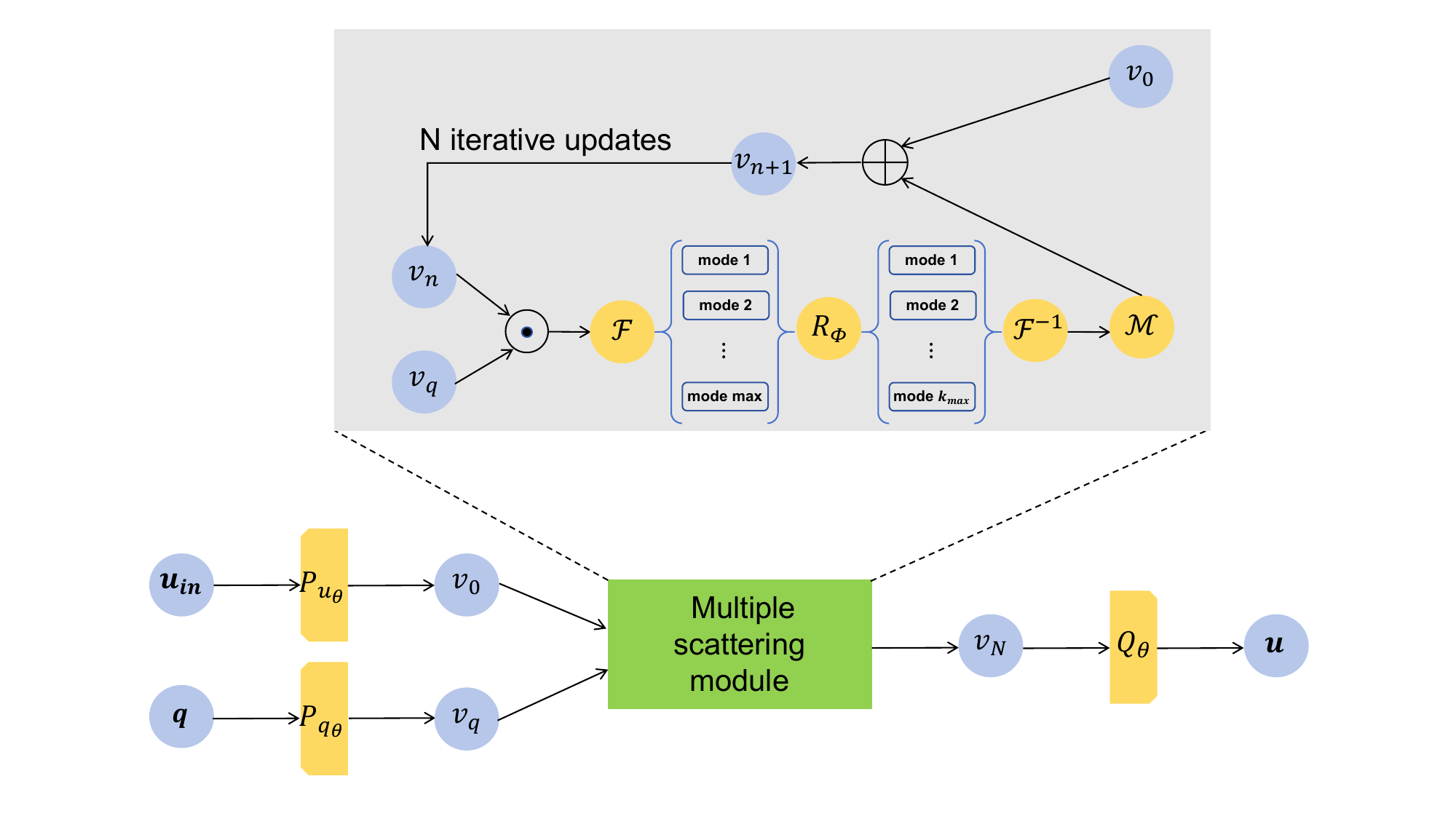}    
\caption{Network architecture of our Multiple Scattering Neural Operator (MSNO).}    
\label{fig:MSNO}
\end{figure}

The proposed MSNO, illustrated in Figure~\ref{fig:MSNO}, approximates the total-field solution of the Helmholtz equation through an architecture inspired by the Lippmann--Schwinger formulation~\eqref{eq:LS}. Starting from the incident-field embedding $v_0$, the model performs $N$ recursive updates conditioned on the fixed medium representation $v_q$. The final latent state $v_N$ is then decoded to obtain the predicted total field $u_\theta$. The update parameters are shared across iterations. This construction follows the recursive structure of the Born iteration in Eq.~\eqref{eq:born_iter_formal}, but the learned updates are not identified with individual physical scattering orders and do not automatically inherit the convergence properties of the classical iteration.

\textbf{Spectral parametrization of the scattering operator.}
We define a learnable scattering operator $\mathcal{S}_{\theta}$ acting on the latent wavefield and medium representations. The wavefield features are updated according to
\begin{equation}
v_{n+1}
=
v_0+\mathcal{S}_{\theta}(v_n,v_q),
\qquad n=0,1,\ldots,N-1,
\label{eq:latent_recursion}
\end{equation}
where $v_n$ and $v_{n+1}$ denote the latent wavefield states before and after the $(n+1)$-th update, respectively. The incident-field embedding $v_0$ serves as both the initial state and a fixed additive reference at every update. The operator $\mathcal{S}_{\theta}$ is a learned latent-space analogue of the physical scattering operator $\mathcal{S}_q$, with medium dependence supplied through $v_q$. Unlike $\mathcal{S}_q$, which is linear in the wavefield for fixed $q$, $\mathcal{S}_{\theta}$ incorporates nonlinear transformations.
The same operator $\mathcal{S}_{\theta}$ and its parameters are shared across all recursive updates. For a given input sample, $v_0$ and $v_q$ remain fixed, while $v_n$ evolves. Thus, increasing the recursion depth does not introduce additional trainable parameters in the update block, although it increases the computational cost.

For the homogeneous background considered here, the Green's function $\Phi(x,y)$ depends only on $|x-y|$. The physical scattering integral therefore has a convolution structure when the product $q(x)u(x)$ is extended by zero outside $D$. Motivated by the convolution theorem \cite{grafakos2014classical}, we parameterize $\mathcal{S}_{\theta}$ through a learned spectral interaction followed by a pointwise nonlinear transformation:
\begin{equation}
\mathcal{S}_{\theta}(v_n,v_q)(x)
:=
\mathcal{M}\left(
\mathcal{F}^{-1}
\Bigl(
R_{\Phi}\cdot
\mathcal{F}\bigl(v_q\odot v_n\bigr)
\Bigr)(x)
\right),
\label{eq:spectral_para}
\end{equation}
where $\mathcal{F}$ and $\mathcal{F}^{-1}$ denote the discrete Fourier transform and its inverse on the computational grid. The symbol $\odot$ denotes pointwise multiplication of corresponding feature channels, modeling medium--wavefield interactions in latent space. At each retained Fourier mode, $R_{\Phi}$ acts as a complex-valued matrix on the transformed feature vector, mixing channels without coupling distinct Fourier modes. This learned multiplier is motivated by the convolution structure of the physical scattering integral; it is not prescribed as the exact Fourier transform of the background Green's function. In particular, the discrete spectral interaction is a learned operation on the computational grid rather than an exact evaluation of the free-space volume potential. The pointwise transition block $\mathcal{M}=\sigma\circ W_1\circ\sigma\circ W_0$
consists of learnable linear transformations $W_0,W_1$ and a nonlinear activation function $\sigma$, applied identically at each spatial location. For fixed $v_q$, the spectral interaction is linear in $v_n$, whereas the complete operator $\mathcal{S}_{\theta}$ is generally nonlinear because of $\mathcal{M}$.

In practice, the spectral interaction is restricted to a finite set of retained Fourier modes:
{
\[
\mathcal{K}
=
\left\{
\boldsymbol{\xi}\in\mathbb{Z}^{d}:
|\xi_j|\le K_{\max,j},\quad j=1,\ldots,d
\right\},
\]
}where $d$ is the spatial dimension, with $d=2$ in the present setting, and $\boldsymbol{\xi}$ denotes the discrete Fourier-mode index. The truncation indices $K_{\max,j}$ are chosen so that $\mathcal{K}$ lies within the modes resolved by the computational grid. We denote the number of retained modes by $|\mathcal{K}|$ and set the multiplier to zero outside $\mathcal{K}$.
For each $\boldsymbol{\xi}\in\mathcal{K}$, the matrix $R_{\Phi}(\boldsymbol{\xi})\in\mathbb{C}^{d_v\times d_v}$ mixes the latent feature channels. Collectively, these matrices form the tensor $R_{\Phi}\in\mathbb{C}^{|\mathcal{K}|\times d_v\times d_v}$.

The discrete Fourier transforms are evaluated using the Fast Fourier Transform (FFT), with a cost of
$\mathcal{O}(N_{\mathrm{grid}}\log N_{\mathrm{grid}})$
per feature channel, where $N_{\mathrm{grid}}$ is the number of spatial grid points. Dense channel mixing over the retained modes costs
$\mathcal{O}(|\mathcal{K}|d_v^2)$.
Mode truncation reduces the number of spectral parameters and the cost of channel mixing, while the FFT cost still depends on the full grid size. It restricts the frequencies processed by the spectral multiplier without distinguishing noise from physically meaningful high-frequency content. The subsequent nonlinear transformation can generate components outside the retained spectral band.

Combining the latent recursion~\eqref{eq:latent_recursion} with the spectral parametrization~\eqref{eq:spectral_para}, we obtain
\begin{equation}
v_{n+1}(x)
=
v_0(x)
+
\mathcal{M}\left(
\mathcal{F}^{-1}
\Bigl(
R_{\Phi}\cdot\mathcal{F}(v_q\odot v_n)
\Bigr)(x)
\right),
\qquad
n=0,1,\ldots,N-1,
\label{eq:MSNO_full_update}
\end{equation}
where $x$ ranges over the computational grid points in $D$. Both $R_{\Phi}$ and the parameters of $\mathcal{M}$ are shared across recursive updates. The block $\mathcal{M}$ transforms the nonlocal features obtained from the medium-modulated wavefield, while $v_0$ provides a fixed incident-field reference. After $N$ recursive updates, the final latent state $v_N$ is decoded to obtain the predicted total field $u_\theta$. By combining Fourier-domain interactions with shared recursive updates, MSNO is designed to represent nonlocal wavefield dependencies and repeated interactions with the medium. Its nonlinear latent recursion is inspired by a finite number of Born iterations rather than an exact truncated Born series.

\subsection{Loss function}
Let $\{(q_j,u_{\mathrm{in},j},u_j)\}_{j=1}^{m}$ denote the training dataset, and let $u_{\theta,j}$ be the predicted total field for the $j$-th sample. The data loss is
\[
\mathcal{L}_{\mathrm{data}}
=
\frac{1}{m}\sum_{j=1}^{m}
\left\|u_{\theta,j}-u_j\right\|_2^2,
\]
where the squared norm sums the squared magnitudes of the complex-valued errors over the spatial grid, equivalently summing the squared errors in the real and imaginary channels.

\textbf{Autoencoder loss.}
The incident-field reconstruction consistency described in Section~\ref{sec3} is encouraged by the auxiliary loss
\[
\mathcal{L}_{\mathrm{ae}}
=
\frac{1}{m}\sum_{j=1}^{m}
\left\|
Q_\theta(v_{0,j})
-
\big[
\operatorname{Re}\{u_{\mathrm{in},j}\},\
\operatorname{Im}\{u_{\mathrm{in},j}\}
\big]
\right\|_2^2,
\]
where $v_{0,j}$ is the initial incident-field embedding, and the norm is evaluated over the spatial grid and the two reconstruction channels.

\textbf{Total loss.}
The training objective combines the prediction and reconstruction losses:
\[
\mathcal{L}_{\mathrm{total}}
=
\mathcal{L}_{\mathrm{data}}
+
\lambda\mathcal{L}_{\mathrm{ae}},
\]
where $\lambda>0$ weights the auxiliary loss and is set to $0.2$ in the numerical experiments.

\section{Multi-scale Multiple Scattering Neural Operator}
\label{sec:ms_msno}

Previous studies \cite{rahaman2019spectral,xu2019frequency} have identified spectral bias in neural-network function approximation, whereby lower-frequency components of a target function are often learned faster than higher-frequency components. This tendency depends on the architecture, training procedure, and target function. In Helmholtz wave propagation and inverse scattering, lower spatial frequencies describe broad variations, while higher spatial frequencies contribute to rapid wavefield oscillations and fine-scale features. Representing sharp interfaces requires a broad range of spatial frequencies, and phase information is distributed across the spectrum. Preferential learning of lower-frequency components may therefore hinder the approximation of oscillatory wavefields and fine-scale medium structures. Related training difficulties have also been analyzed for physics-informed neural networks \cite{wang2022and}. These observations motivate a multiscale extension of MSNO that combines scale-conditioned representations to improve wavefield approximation across spatial scales.

\subsection{Multi-scale strategy}

\begin{figure}[t]    
\centering    
\includegraphics[width=0.8\linewidth]{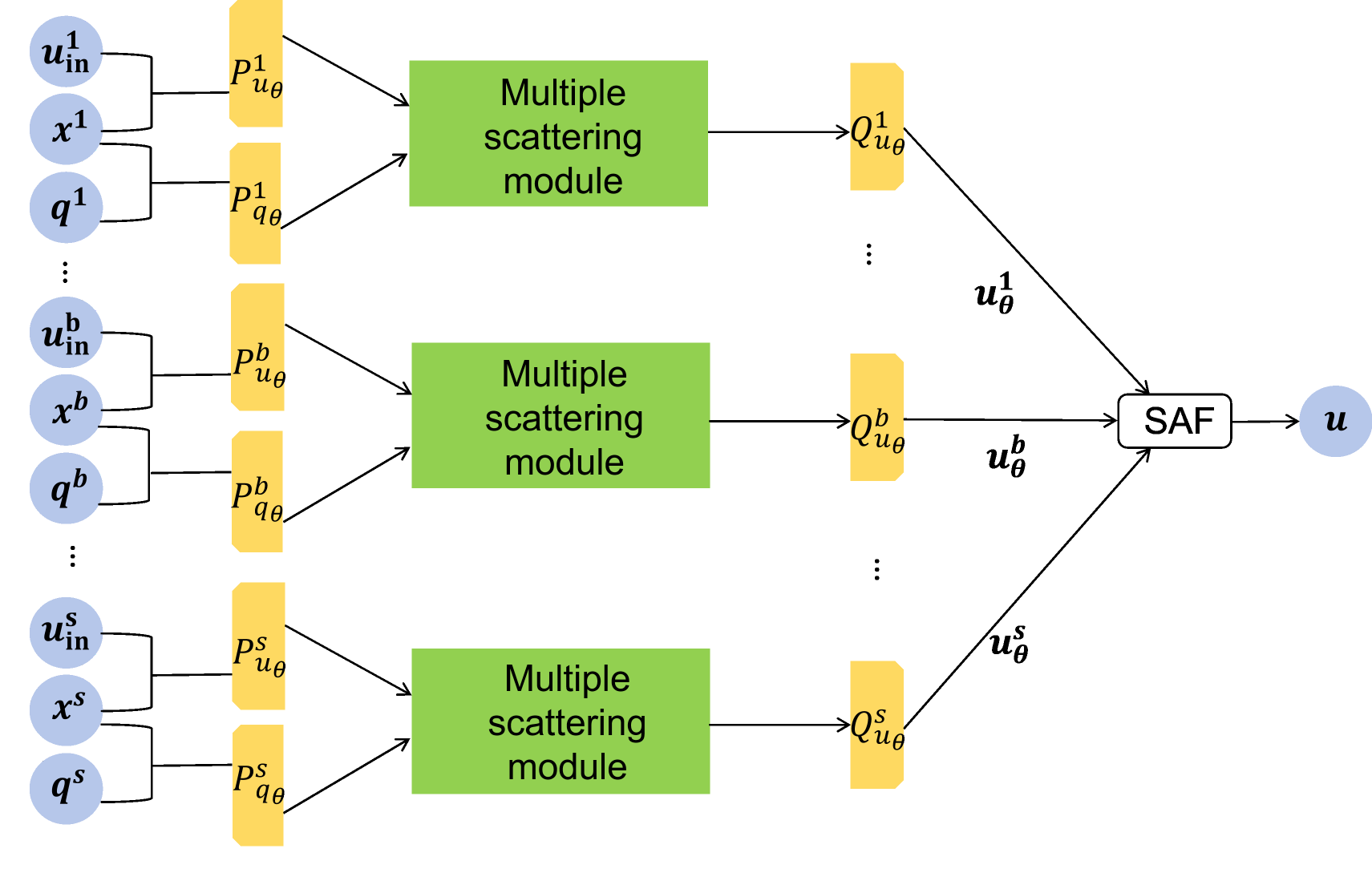}    
\caption{Architecture of the multi-scale MSNO, which comprises a total of $s$ scales.}    
\label{fig:MS-MSNO}
\end{figure}

Inspired by the multiscale strategies in \cite{mscalefno,liu2020multi}, we introduce the Multi-Scale Multiple Scattering Neural Operator (MS-MSNO), illustrated in Figure~\ref{fig:MS-MSNO}. The framework retains the Born-inspired recursive structure of MSNO and introduces scale conditioning through parallel branches. We define the scale factors as
\[
\mathcal{A}_s
=
\{\alpha_1,\alpha_2,\ldots,\alpha_s\},
\qquad
1=\alpha_1<\alpha_2<\cdots<\alpha_s,
\]
where $s$ denotes the number of branches. For branch $b=1,\ldots,s$, the scaled inputs are
\[
x^{(b)}=\alpha_b x,\qquad
q^{(b)}(x)=\alpha_b q(x),\qquad
u_{\mathrm{in}}^{(b)}(x)=\alpha_b u_{\mathrm{in}}(x),
\]
where $x^{(b)}$ is the scaled coordinate supplied to the embedding networks, whereas $q^{(b)}$ and $u_{\mathrm{in}}^{(b)}$ are amplitude-scaled fields evaluated at the original grid points. The superscript $(b)$ denotes the branch index. These transformations condition the latent representations without resampling the fields or changing the computational grid. They do not define separate physical scattering problems; all branches contribute to predicting the same unscaled total field.

The coordinate scaling is motivated by the Fourier scaling identity \cite{grafakos2014classical}: for any compactly supported $q\in L^\infty(\mathbb{R}^2;\mathbb{R})$ and $\alpha>0$,
\[
\widehat{q(\cdot/\alpha)}(\boldsymbol{\xi})
=
\alpha^2\widehat{q}(\alpha\boldsymbol{\xi}),
\]
where, in this identity, the hat denotes the continuous Fourier transform and $\boldsymbol{\xi}\in\mathbb{R}^2$ is a continuous frequency variable. For $\alpha>1$, spatial dilation therefore compresses the corresponding frequency scale by a factor of $\alpha$. This identity provides motivation for coordinate scaling, but supplying scaled coordinates to the embedding networks does not itself implement a spatial dilation of the input fields. Likewise, amplitude scaling changes Fourier magnitudes without shifting frequency support.

In the proposed construction, all branch features remain defined on the original computational grid. Since each branch uses the same retained Fourier-mode set $\mathcal{K}$, input scaling does not enlarge the set of modes processed by the spectral multipliers. Instead, the branches learn nonlinear representations conditioned on different input scalings, whose combination is intended to improve approximation across spatial scales. The first branch, with $\alpha_1=1$, uses the unscaled MSNO inputs. For each branch $b=1,\ldots,s$, the incident field and medium contrast are embedded as
\[
v_0^{(b)}(x)
=
P_{u_\theta}^{(b)}\!\left(
x^{(b)},\operatorname{Re}\{u_{\mathrm{in}}^{(b)}(x)\},
\operatorname{Im}\{u_{\mathrm{in}}^{(b)}(x)\}
\right),
\]
and
\[
v_q^{(b)}(x)
=
P_{q_\theta}^{(b)}\!\left(x^{(b)},q^{(b)}(x)\right),
\]
respectively. The latent wavefield is then updated using the MSNO recursion:
\[
v_{n+1}^{(b)}(x)
=
v_0^{(b)}(x)
+
\mathcal{M}^{(b)}\left(
\mathcal{F}^{-1}
\Bigl(
R_{\Phi}^{(b)}\cdot
\mathcal{F}\bigl(v_q^{(b)}\odot v_n^{(b)}\bigr)
\Bigr)(x)
\right),
\quad n=0,\ldots,N-1,
\]
where $v_q^{(b)}$ modulates the latent wavefield through pointwise multiplication, while $v_0^{(b)}$ provides the initial state and a fixed additive reference. The parameters of $R_{\Phi}^{(b)}$ and $\mathcal{M}^{(b)}$ are specific to each branch and shared across its recursive updates. For a given input sample, $v_q^{(b)}$ and $v_0^{(b)}$ remain fixed while $v_n^{(b)}$ evolves.

After $N$ recursive updates, each branch produces a total-field prediction on the original spatial grid:
\[
\big[
\operatorname{Re}\{u_{\theta}^{(b)}(x)\},\
\operatorname{Im}\{u_{\theta}^{(b)}(x)\}
\big]
=
Q_{\theta}^{(b)}\!\left(v_N^{(b)}(x)\right),
\qquad b=1,\ldots,s,
\]
where the brackets denote concatenation along the channel dimension. All branch outputs use the same spatial coordinates and field representation as the unscaled target and are subsequently combined by the fusion module.

\subsection{Spectrally adaptive fusion}

For each branch $b=1,\ldots,s$, we compute the two-dimensional discrete Fourier transform of the predicted complex-valued total field:
\[
\widehat{u}_{\theta}^{(b)}(\boldsymbol{\xi})
=
\mathcal{F}\bigl(u_{\theta}^{(b)}\bigr)(\boldsymbol{\xi}).
\]
Let $\Xi$ denote the discrete Fourier grid. We partition the analyzed frequency range into nonempty, disjoint annular bands:
\begin{equation}
A_j
=
\left\{
\boldsymbol{\xi}\in\Xi:
r_j\leq \|\boldsymbol{\xi}\|_2<r_{j+1}
\right\},
\quad
r_{j+1}-r_j=\Delta r>0,
\quad j=1,\ldots,J,
\label{eq:band}
\end{equation}
where $J$ is the number of radial-frequency bands, and $r_j$ and $r_{j+1}$ are the inner and outer radii of band $A_j$, respectively. The mean spectral energy of branch $b$ within $A_j$ is
\[
E_b(j)
=
\frac{1}{|A_j|}
\sum_{\boldsymbol{\xi}\in A_j}
\left|\widehat{u}_{\theta}^{(b)}(\boldsymbol{\xi})\right|^2,
\quad
j=1,\ldots,J,
\]
where $|A_j|$ denotes the number of Fourier grid points in $A_j$, and the absolute value denotes the modulus of a complex Fourier coefficient. The logarithmic spectral response is defined as
\begin{equation}
D_b^{\log}(j)
=
\log_{10}\bigl(E_b(j)+\varepsilon\bigr),
\quad b=1,\ldots,s,
\quad j=1,\ldots,J,
\label{eq:red}
\end{equation}
where $\varepsilon>0$ is a small numerical offset that prevents taking the logarithm of zero.
The logarithmic transformation compresses the dynamic range of the spectral energies while preserving their ordering across branches within each band. Combined with Softmax normalization, it produces weights proportional to a power of the shifted energies rather than an exponential of the original energies, as shown below. This formulation moderates sensitivity to large energy values, with the temperature parameter controlling the concentration of the weights.

For each band $A_j$, the spectral responses are normalized across branches using Softmax:
\[
w_b(j)
=
\frac{\exp\bigl(D_b^{\log}(j)/T\bigr)}
{\sum_{\ell=1}^{s}\exp\bigl(D_\ell^{\log}(j)/T\bigr)},
\qquad
\sum_{b=1}^{s}w_b(j)=1,
\]
where $T>0$ is the temperature parameter. Larger values of $T$ yield more uniform weights, whereas smaller values favor branches with larger spectral responses. Substituting the definition of $D_b^{\log}(j)$ gives
\[
w_b(j)
=
\frac{\bigl(E_b(j)+\varepsilon\bigr)^{\beta}}
{\sum_{\ell=1}^{s}\bigl(E_\ell(j)+\varepsilon\bigr)^{\beta}},
\qquad
\beta=\frac{1}{T\ln 10}.
\]
Thus, the weights are proportional to the $\beta$-th powers of the shifted spectral energies. The branch-level fusion coefficients are obtained by averaging the weights over all analyzed bands:
\[
\gamma_b
=
\frac{1}{J}\sum_{j=1}^{J}w_b(j),
\qquad b=1,\ldots,s.
\]
Each band contributes equally to this average, and the coefficients satisfy
$\gamma_b\geq0$ and $\sum_{b=1}^{s}\gamma_b=1$.
The fused total-field prediction is
\begin{equation}
u_\theta(x)
=
\sum_{b=1}^{s}\gamma_b\,u_{\theta}^{(b)}(x).
\label{eq:SAF_fusion}
\end{equation}
Each coefficient $\gamma_b$ is applied uniformly over the spatial grid to both the real and imaginary components of the corresponding branch output. Thus, band-wise spectral statistics determine global branch weights; fusion is not performed separately within each Fourier band. These coefficients reflect relative predicted spectral energies rather than direct estimates of branch accuracy. The fusion coefficients $\gamma_b$ are sample-dependent quantities, not fixed trainable parameters. During training and testing, they are recomputed separately for each sample from the current branch predictions in every forward pass. During inversion, the trained network parameters remain fixed, while the fusion coefficients are recalculated whenever the forward model is evaluated at the current medium estimate. SAF therefore adapts the branch contributions to the spectral characteristics of the current predictions.

\section{Numerical Results}\label{sec5}

\subsection{Experimental setting}
We conduct numerical experiments to evaluate the forward prediction and inverse reconstruction performance of MSNO and MS-MSNO relative to representative neural-operator baselines. Unless otherwise specified, all single-scale models use a latent channel dimension of $d_v=50$. MSNO and BFNO use $N=5$ recursive updates, while FNO uses five Fourier layers. Each branch of the multiscale architectures uses the same channel dimension and update or layer count as its single-scale counterpart. All models are trained using the Adam optimizer for 1000 epochs, with an initial learning rate of $10^{-3}$ that is halved every 230 epochs. The batch size is 32. All experiments are performed on a single NVIDIA GPU with 24~GB of memory.

\textbf{Baselines.} We compare MSNO with the following neural-operator baselines:
\begin{itemize}
    \item \textbf{FNO} \cite{FNO}:
    The Fourier Neural Operator learns the forward solution map using integral operators parameterized in the Fourier domain. We use a channel dimension of $d_v=50$ and five Fourier layers.

    \item \textbf{BFNO} \cite{BFNO}:
    The Born Fourier Neural Operator incorporates a Born-inspired recursive structure into a Fourier neural-operator architecture. We use the same channel dimension and number of recursive updates as MSNO.
\end{itemize}
These settings align the latent widths and update or layer counts, but do not imply identical parameter counts or computational costs.

\textbf{Datasets.} We evaluate the models on two datasets representing different physical settings and structural characteristics: a synthetic electromagnetic dataset with layered media and embedded inclusions, and an acoustic breast-phantom dataset with heterogeneous sound-speed distributions.
\begin{itemize}
    \item \textbf{Synthetic multilayer dataset} \cite{BFNO}:
    The dataset is generated using the finite-difference time-domain simulator gprMax and contains 400 samples, of which 350 are used for training and 50 for testing. The computational domain is a uniformly discretized two-dimensional region of size
    $0.5\,\mathrm{m}\times0.5\,\mathrm{m}$,
    containing three background layers. The thickness of the second layer is sampled as
    $d_2\sim\mathcal{U}(0.1\,\mathrm{m},0.3\,\mathrm{m})$,
    and the thickness of the first layer is
    $d_1=0.35\,\mathrm{m}-d_2$.
    The relative permittivities of the first and second layers are sampled from
    $\epsilon_1\sim\mathcal{U}(3,5)$ and
    $\epsilon_2\sim\mathcal{U}(5,10)$,
    respectively. Two cylindrical inclusions, represented by circular cross-sections in the two-dimensional domain, have radii sampled from
    $\mathcal{U}(0.03\,\mathrm{m},0.06\,\mathrm{m})$.
    Their relative permittivities are
    $\epsilon_{c_1}=1$ and
    $\epsilon_{c_2}\sim\mathcal{U}(3,10)$.
    The simulated time-domain data are transformed into the frequency domain, and data at 50 discrete frequencies between
    $0.5$ and $1.5\,\mathrm{GHz}$ are retained.
    The corresponding relative-permittivity maps serve as reference medium parameters.

    \item \textbf{Breast-phantom dataset} \cite{zeng2025openbreastus}:
    We use 500 heterogeneous breast phantoms from OpenBreastUS, with 450 samples for training and 50 for testing. The original
    $480\times480$ sound-speed maps are downsampled to
    $160\times160$ using bicubic interpolation and represented on the normalized coordinate domain
    $\Omega=[0,1]^2$.
    The ultrasound computed tomography (USCT) acquisition geometry consists of a circular array with a diameter of
    $220\,\mathrm{mm}$ and 256 transducers.
    Eight uniformly spaced transducer positions are selected as sources, with an angular separation of $2\pi/8$ radians.
    The sound-speed maps serve as reference medium parameters.
\end{itemize}
For all experiments, the input and output fields are standardized using unit Gaussian normalization to facilitate optimization \cite{huang2023normalization}.

\subsection{Evaluation on forward simulation}

\begin{figure}[t]    
\centering    
\includegraphics[width=0.95\linewidth]{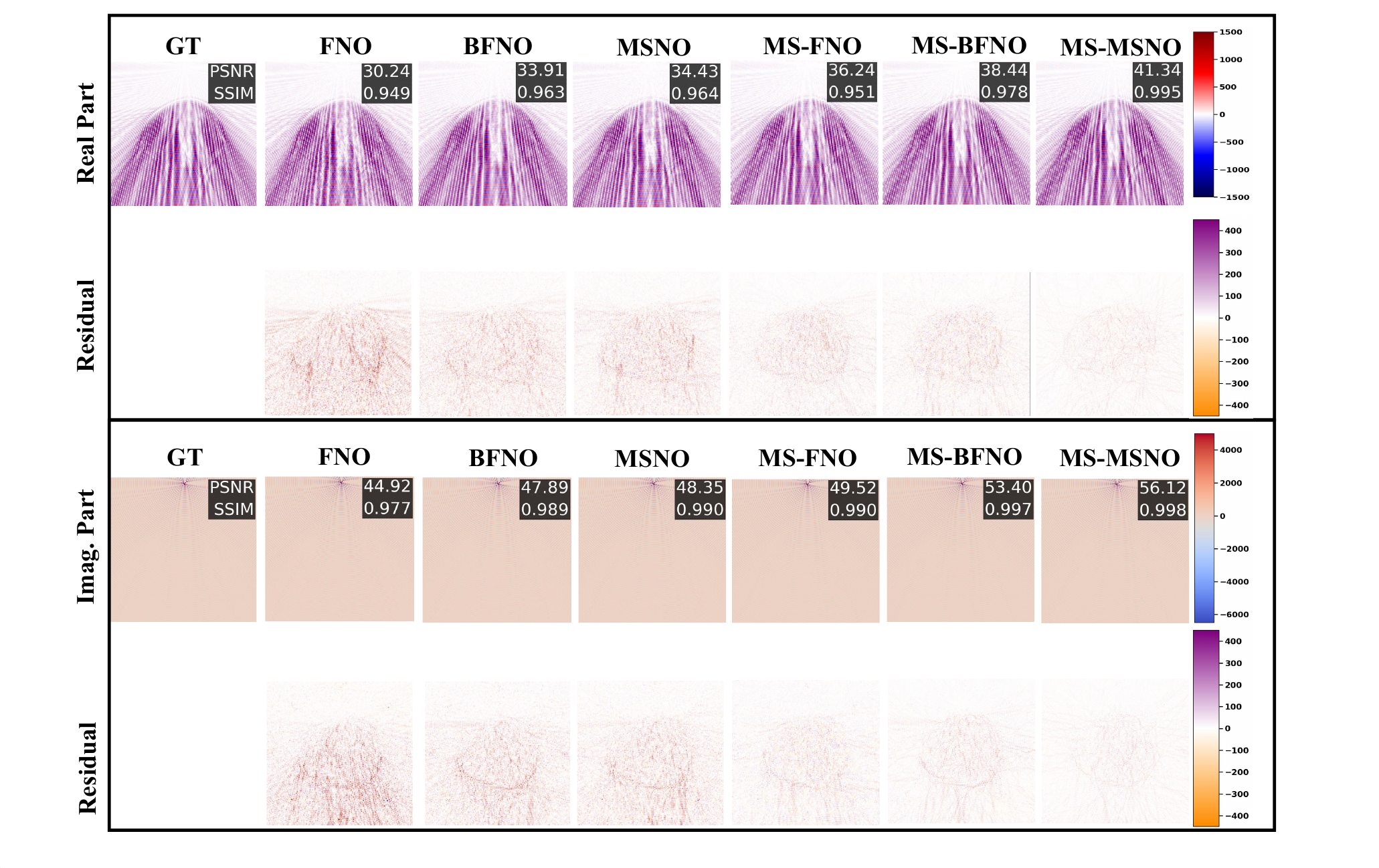}    
\caption{Comparison of simulated total wavefields obtained by different benchmark methods. The real and imaginary components of the wavefields are shown separately, with the real part displayed on the top and the imaginary part on the bottom, enabling a clear visual assessment of the wave propagation characteristics captured by each method.}    
\label{fig:forward}
\end{figure}

We evaluate MSNO and MS-MSNO as approximations of the forward solution map associated with the scattering problem~\eqref{eq:scattering_system}. Given the medium contrast $q(x)$ and the incident field $u_{\mathrm{in}}$, the models predict the total acoustic pressure field $u=u_{\mathrm{in}}+u_{\mathrm{sca}}$. Accurate forward prediction is important for gradient-based inversion in USCT, since errors in the learned forward operator can affect the data-misfit objective and the resulting medium reconstruction. Figure~\ref{fig:forward} presents a visual comparison of the predicted wavefields, while Table~\ref{tab:wave} reports the relative $L^2$ error, PSNR, and SSIM for their real and imaginary components. Among the models compared, MS-MSNO achieves the lowest relative $L^2$ errors and the highest PSNR and SSIM values for both components. For the real component, it obtains a relative $L^2$ error of $6.09\%$, a PSNR of $41.34\,\mathrm{dB}$, and an SSIM of $0.995$. For the imaginary component, the corresponding values are $3.53\%$, $56.12\,\mathrm{dB}$, and $0.998$, respectively.

\begin{table}[h]
\centering
\caption{Quantitative comparison of simulation performance across different neural operator frameworks. The best results are highlighted in bold.}
\begin{tabular}{lcccccc}
\toprule
\multirow{2}{*}{\textbf{Method}} & \multicolumn{3}{c}{\textbf{Real Part}} & \multicolumn{3}{c}{\textbf{Imaginary Part}} \\
\cmidrule(lr){2-4} \cmidrule(lr){5-7}
& Rel $L^2$ ($\downarrow$) & PSNR ($\uparrow$) & SSIM ($\uparrow$) & Rel $L^2$ ($\downarrow$) & PSNR ($\uparrow$) & SSIM ($\uparrow$) \\
\midrule
FNO      & 22.36\% & 30.24 & 0.949 & 12.81\% & 44.92 & 0.977 \\
BFNO     & 14.67\% & 33.91 & 0.963 & 9.09\%  & 47.89 & 0.989 \\
MSNO     & 13.81\% & 34.43 & 0.964 & 8.63\%  & 48.35 & 0.990 \\
MS-FNO  & 10.13\%  & 36.24 & 0.951 & 6.55\%  & 49.52 & 0.990 \\
MS-BFNO  & 8.07\%  & 38.44 & 0.978 & 4.83\%  & 53.40 & 0.997 \\
\rowcolor{gray!15} 
\textbf{MS-MSNO} & \textbf{6.09\%} & \textbf{41.34} & \textbf{0.995} & \textbf{3.53\%} & \textbf{56.12} & \textbf{0.998} \\
\bottomrule
\end{tabular}
\label{tab:wave}
\end{table}

\begin{figure}[h]    
\centering    
\includegraphics[width=1\linewidth,height=0.5\textheight,keepaspectratio]{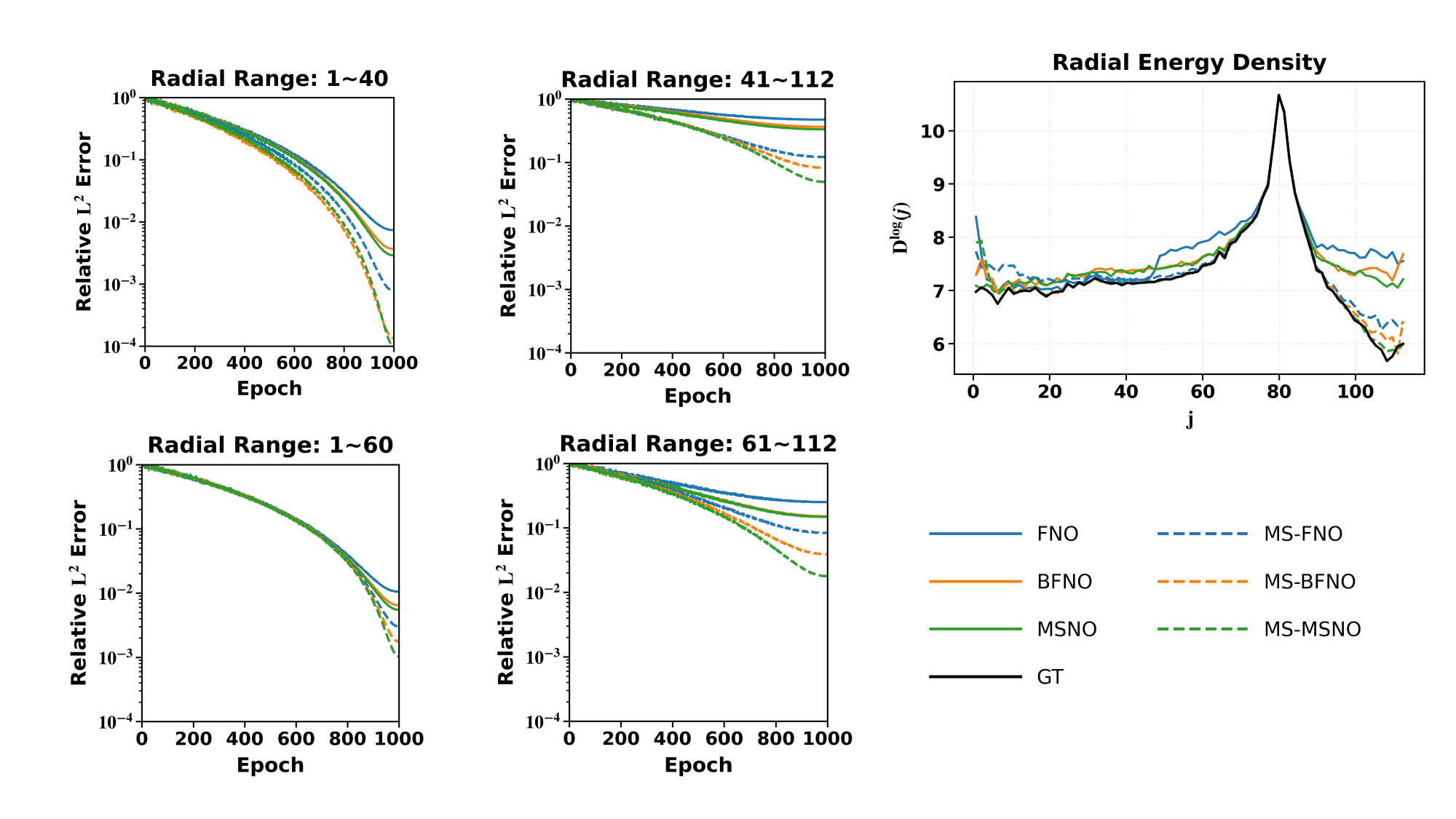}    
\caption{
Frequency-domain error and radial energy density analyses for different neural
operator models. The four error panels show the evolution of the band-limited
relative $L^2$ error during training under two radial-frequency partitions.
The first row corresponds to the cutoff $c=40$, with
$B_{1:40}$ and
$B_{41:112}$, while the second row corresponds to the
cutoff $c=60$, with
$B_{1:60}$ and
$B_{61:112}$.
The radial-frequency partitions are introduced as empirical analysis settings
to examine the sensitivity of the spectral-error comparison to the cutoff
choice. The right panel compares the logarithmic radial energy density
$D^{\log}(j)$ of the predicted wavefields with that of the ground-truth
wavefield. Solid and dashed curves denote the single-scale and multi-scale
model variants, respectively.}    
\label{fig:radial_energy}
\end{figure}

We use radial spectral energy analysis \cite{kanafi_radially_2016} to evaluate the spectral approximation performance of different wavefield prediction models. Based on the annular partition defined in Eq.~\eqref{eq:band}, the logarithmic mean spectral energy within each annulus is computed according to Eq.~\eqref{eq:red}. Figure~\ref{fig:radial_energy} compares the spectral characteristics of the predicted and reference wavefields. For a radial-frequency band $\mathcal{B}_{a:b}:=\left\{
\boldsymbol{\xi}\in\Xi:a\leq \|\boldsymbol{\xi}\|_2<b\right\}$, where $a$ and $b$ index consecutive annular bins, the band-limited relative $L^2$ error is defined as
\[
e_{a:b}
=
\left(
\frac{
\displaystyle\sum_{\boldsymbol{\xi}\in\mathcal{B}_{a:b}}
\left|
\widehat{u}_{\mathrm{pred}}(\boldsymbol{\xi})
-
\widehat{u}_{\mathrm{ref}}(\boldsymbol{\xi})
\right|^2
}{
\displaystyle\sum_{\boldsymbol{\xi}\in\mathcal{B}_{a:b}}
\left|
\widehat{u}_{\mathrm{ref}}(\boldsymbol{\xi})
\right|^2
}
\right)^{1/2},
\]
where $\widehat{u}_{\mathrm{pred}}$ and $\widehat{u}_{\mathrm{ref}}$ are the Fourier transforms of the predicted and reference total fields, respectively, and the reference energy in the selected band is assumed to be nonzero. This error measures discrepancies in the complex Fourier coefficients, complementing the comparison of radial energy profiles.

The four frequency-error panels show the evolution of the band-limited relative $L^2$ error during training for two radial-frequency partitions with cutoff indices $c=40$ and $c=60$. Under the present discretization, the analyzed radial spectrum comprises 112 bins. The resulting partitions are $\mathcal{B}_{1:40}$ and $\mathcal{B}_{41:112}$ for $c=40$, and $\mathcal{B}_{1:60}$ and $\mathcal{B}_{61:112}$ for $c=60$. These cutoffs are chosen empirically for spectral-error analysis and do not represent intrinsic physical frequency thresholds. The two partitions are used to assess the sensitivity of the model comparison to the cutoff choice.

For both partitions, the models exhibit broadly similar error-decay trends in the lower-frequency bands, while their differences are more pronounced in the upper-frequency bands. In particular, MS-MSNO achieves the lowest upper-frequency error during the later stages of training in both $\mathcal{B}_{41:112}$ and $\mathcal{B}_{61:112}$. Thus, its observed advantage is maintained across the two tested cutoff choices. These results support improved approximation of the upper-frequency wavefield components in the reported experiments, consistent with the motivation for introducing multiscale representations to address spectral bias.

The lower panel compares the logarithmic mean spectral energy $D^{\log}(j)$ of the predicted wavefields with that of the reference field. FNO overestimates the spectral energy at intermediate-to-high radial frequencies, whereas BFNO and MSNO show closer agreement with the reference profile, although discrepancies remain. Compared with their single-scale counterparts, the multiscale variants better follow the reference profile, particularly near the dominant energy peak and in the high-frequency decay region. Among the methods compared, MS-MSNO shows the closest overall agreement in this plot. Together with the band-limited relative $L^2$ errors, these results support the effectiveness of the proposed architecture in representing the spectral content of the wavefield. The radial energy profiles alone, however, do not establish phase accuracy or preservation of fine-scale spatial structures.

\subsection{Evaluation on inverse medium problems}
The inverse medium problem aims to reconstruct the unknown medium contrast $q(x)$ from measured total-field data under prescribed incident excitations. In this work, we evaluate MSNO and MS-MSNO in dielectric and acoustic imaging, focusing on their ability to recover spatially heterogeneous medium properties from measurements at receiver locations. The reconstruction is formulated as the nonlinear least-squares problem in Eq.~\eqref{eq:inverse_problem}, with the forward solution operator replaced by the trained neural surrogate. During inversion, the network parameters remain fixed, while the medium estimate is updated using gradients of the data-misfit objective computed by automatic differentiation through the surrogate. This approach avoids separate forward and adjoint PDE solves at each inversion iteration. The resulting reconstruction accuracy depends on both the approximation quality of the learned forward map and the optimization procedure.

\begin{figure}[htbp]
\centering
\setlength{\tabcolsep}{2pt}
\renewcommand{\arraystretch}{1.05}

\begin{tabular}{c c c c}
\textbf{GT} &
\textbf{FNO} &
\textbf{BFNO} &
\textbf{MSNO} 
\\
\includegraphics[width=0.20\textwidth]{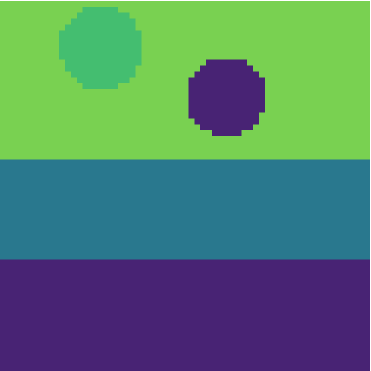} &
\includegraphics[width=0.20\textwidth]{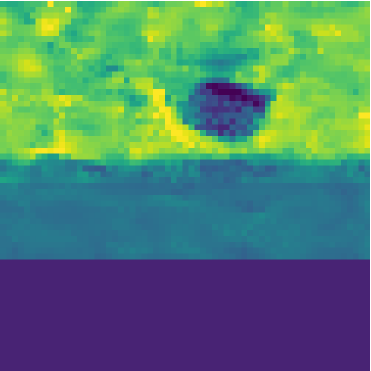} &
\includegraphics[width=0.20\textwidth]{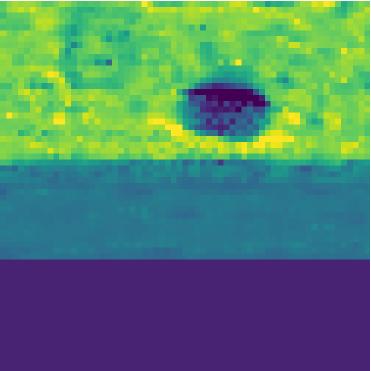} &
\includegraphics[width=0.20\textwidth]{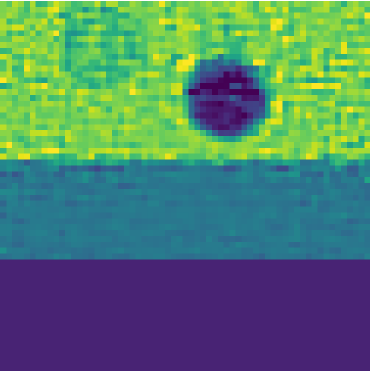} 
\\
\vspace{-0.6cm}
\includegraphics[width=0.20\textwidth]{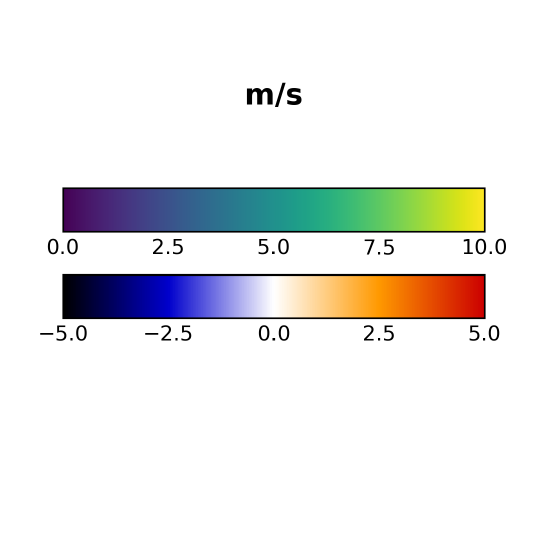}&
\includegraphics[width=0.20\textwidth]{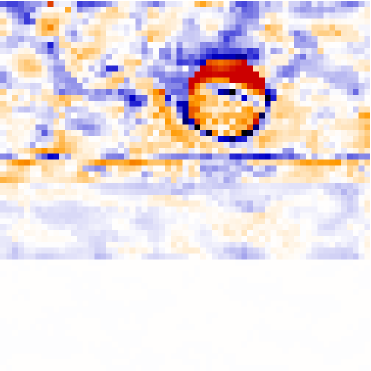} &
\includegraphics[width=0.20\textwidth]{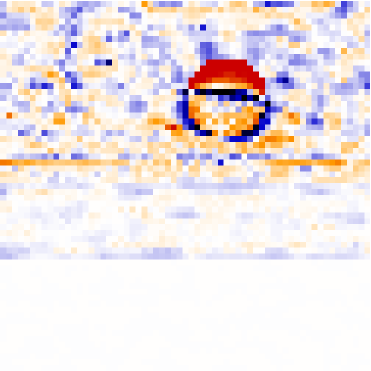} &
\includegraphics[width=0.20\textwidth]{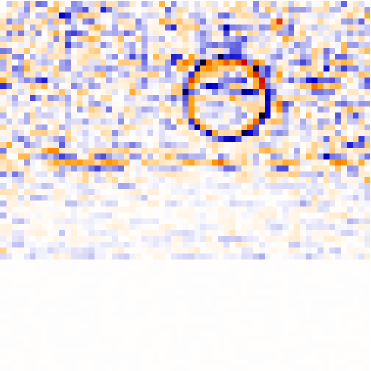} 
\\

\includegraphics[width=0.20\textwidth]{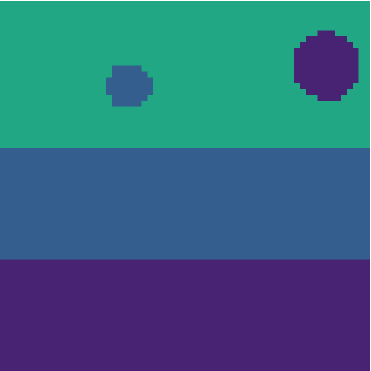} &
\includegraphics[width=0.20\textwidth]{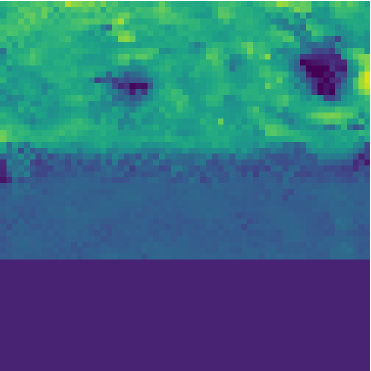} &
\includegraphics[width=0.20\textwidth]{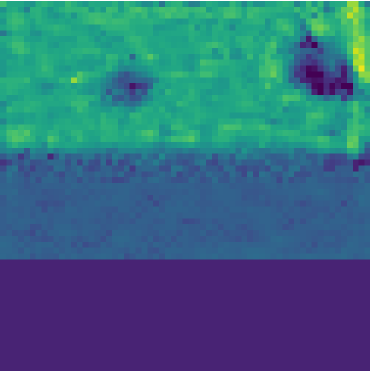} &
\includegraphics[width=0.20\textwidth]{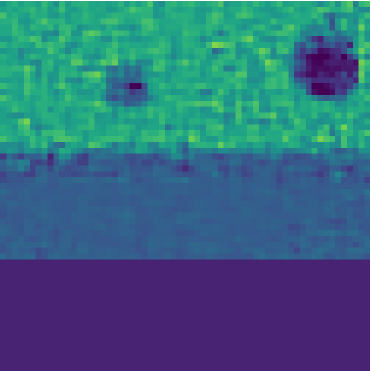} 
\\
\vspace{-0.6cm}
&
\includegraphics[width=0.20\textwidth]{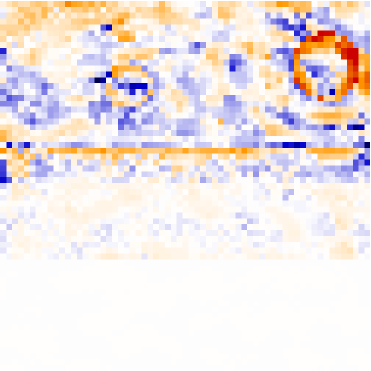} &
\includegraphics[width=0.20\textwidth]{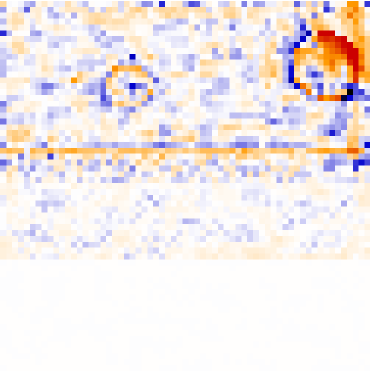} &
\includegraphics[width=0.20\textwidth]{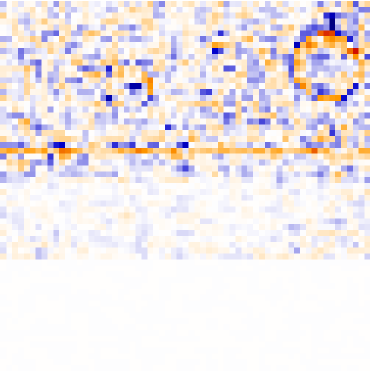} 
\\

\includegraphics[width=0.20\textwidth]{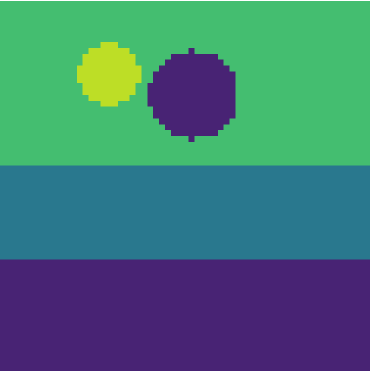} &
\includegraphics[width=0.20\textwidth]{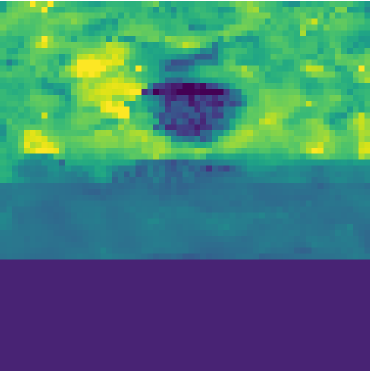} &
\includegraphics[width=0.20\textwidth]{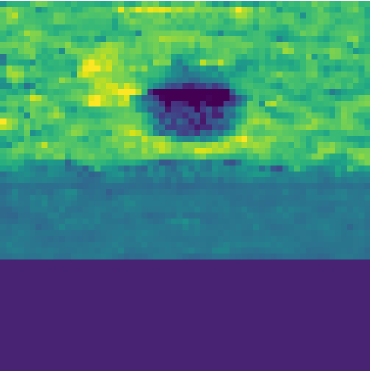} &
\includegraphics[width=0.20\textwidth]{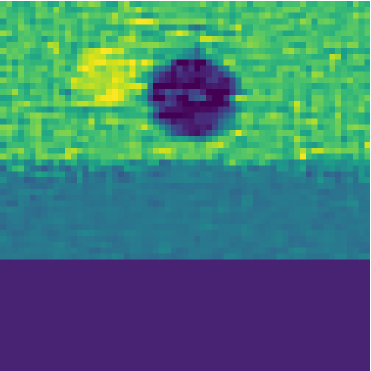} 
\\

&
\includegraphics[width=0.20\textwidth]{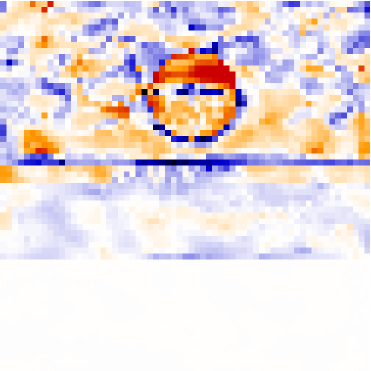} &
\includegraphics[width=0.20\textwidth]{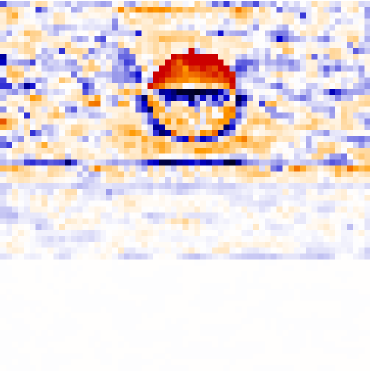} &
\includegraphics[width=0.20\textwidth]{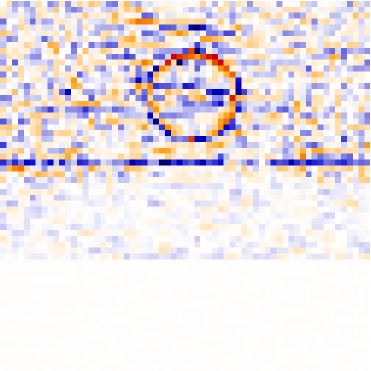} 
\vspace{-0.8cm}
\end{tabular}
\caption{
    Visual comparison of reconstructed medium parameter fields $q(x)$ for three representative samples from the synthetic dataset. Odd-numbered rows display the ground truth (GT) and the reconstruction results for different methods. The even-numbered rows directly below each sample illustrate the corresponding pixel-wise residual maps.
}
\label{fig:gen_data}
\end{figure}

\begin{table}[ht]
\centering
\renewcommand{\arraystretch}{1.2}
\caption{Quantitative comparison of reconstruction performance and computational efficiency for the synthetic multi‑layered dataset. Metrics include relative PSNR, SSIM, inference speed (iterations per second), peak memory consumption, and total trainable parameters.}

\begin{tabular}{c|c|c|c|c|c}
\specialrule{1.5pt}{0pt}{0pt} 
Method  & PSNR ($\uparrow$) & SSIM ($\uparrow$) & Iters(s) ($\downarrow$) & Memo.(GB) ($\downarrow$) & Params.(M) ($\downarrow$) \\
\specialrule{1.5pt}{0pt}{0pt} 
FNO   & 23.14 & 0.5357 & 42 & 9.82 & 32.5 \\
\hline
BFNO  &25.18 & 0.5994 & 30 & 7.3  & 9.2  \\
\hline
\rowcolor{gray!15} 
\textbf{MSNO}  & \textbf{27.32} & \textbf{0.6285} & \textbf{25} & \textbf{7.1} & \textbf{8.7} \\
\specialrule{1.5pt}{0pt}{0pt} 
\end{tabular}
\label{tab:gen_data}
\end{table}

\textbf{Synthetic multilevel experiment.} We first evaluate reconstruction performance on the synthetic multilayer dataset. The qualitative results in Figure~\ref{fig:gen_data} show that MSNO produces smaller reconstruction errors than the FNO and BFNO baselines in the displayed examples. In particular, MSNO more clearly delineates the embedded inclusions and layer interfaces, recovering sharper boundaries and more accurately reproducing the reference medium structures. Table~\ref{tab:gen_data} summarizes the average PSNR and SSIM, together with computational cost and model size. Among the compared methods, MSNO achieves the highest PSNR and SSIM while using fewer trainable parameters and less GPU memory. Parameter sharing across scattering iterations limits the parameter count by reusing the same update block, although the computational cost still depends on the number of recursive updates. These results indicate a favorable balance between reconstruction accuracy and computational requirements for the tested multilayer media.

\begin{figure}[htbp]    
\centering    
\includegraphics[width=1\linewidth]{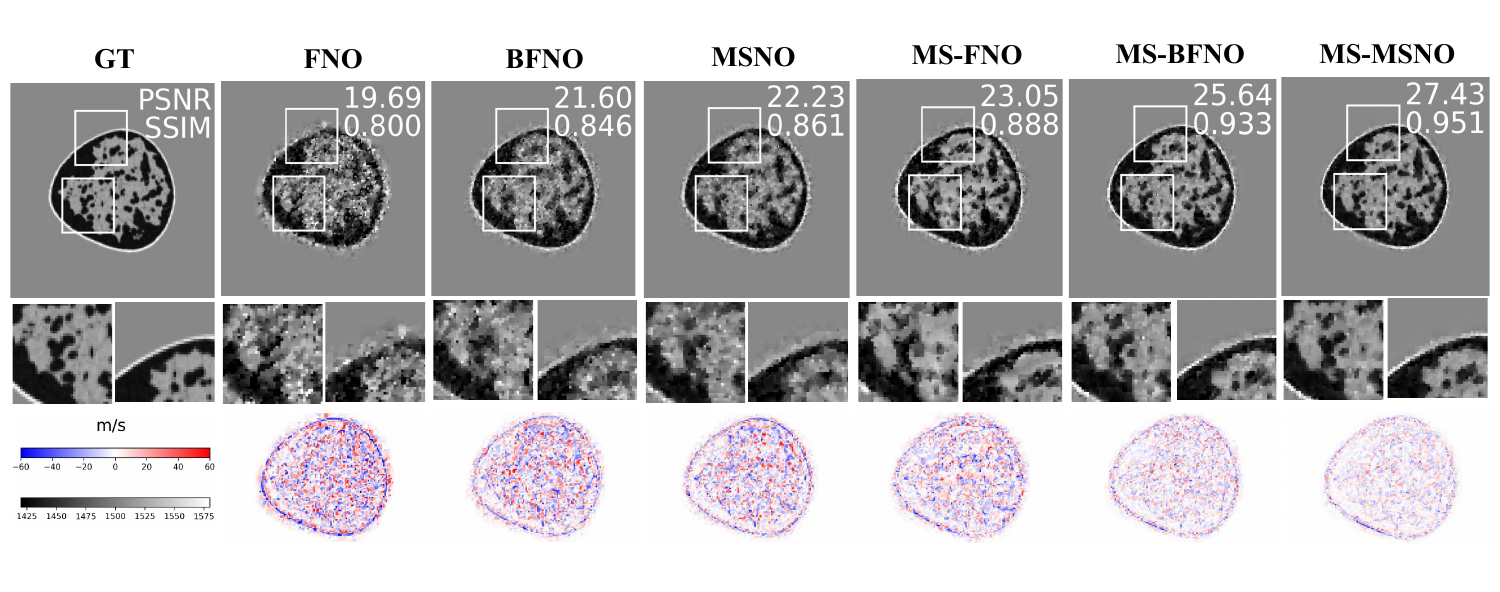}    
\caption{Qualitative comparison of breast phantoms reconstruction results between different models. The rows correspond to the sound-of-speed (SOS), ROI~1, ROI~2, and residual maps.}    
\label{fig:case3}
\end{figure}

\textbf{Breast phantoms experiment.} Breast ultrasound computed tomography (USCT) \cite{fan2022model} is a non-ionizing imaging modality that reconstructs quantitative maps of tissue acoustic properties and may support breast lesion characterization. In the present constant-density acoustic model, the contrast $q(x)$ is related to the sound-speed distribution by $q(x)=c_b^2/c(x)^2-1$, where $c_b$ is the background sound speed. Figure~\ref{fig:case3} presents reconstruction results for a held-out breast slice from the same phantom dataset, using the same spatial resolution and wavenumber setting, with contrasts within the range represented in the training data. For this example, MS-MSNO shows closer agreement with the reference sound-speed map than the compared baselines. The multiscale variants also exhibit improved structural delineation and fewer visible artifacts relative to their single-scale counterparts. In particular, multiscale input conditioning improves the Fourier-based baselines, while MS-MSNO combines this strategy with recursive latent scattering updates and spectral fusion to recover finer anatomical details. These results illustrate the benefits of multiscale modeling for the tested USCT reconstruction task within the training regime; they do not establish generalization to different resolutions, frequencies, or contrast ranges.

\begin{figure}[htp!]    
\centering    
\includegraphics[width=1\linewidth,height=0.8\textheight,keepaspectratio]{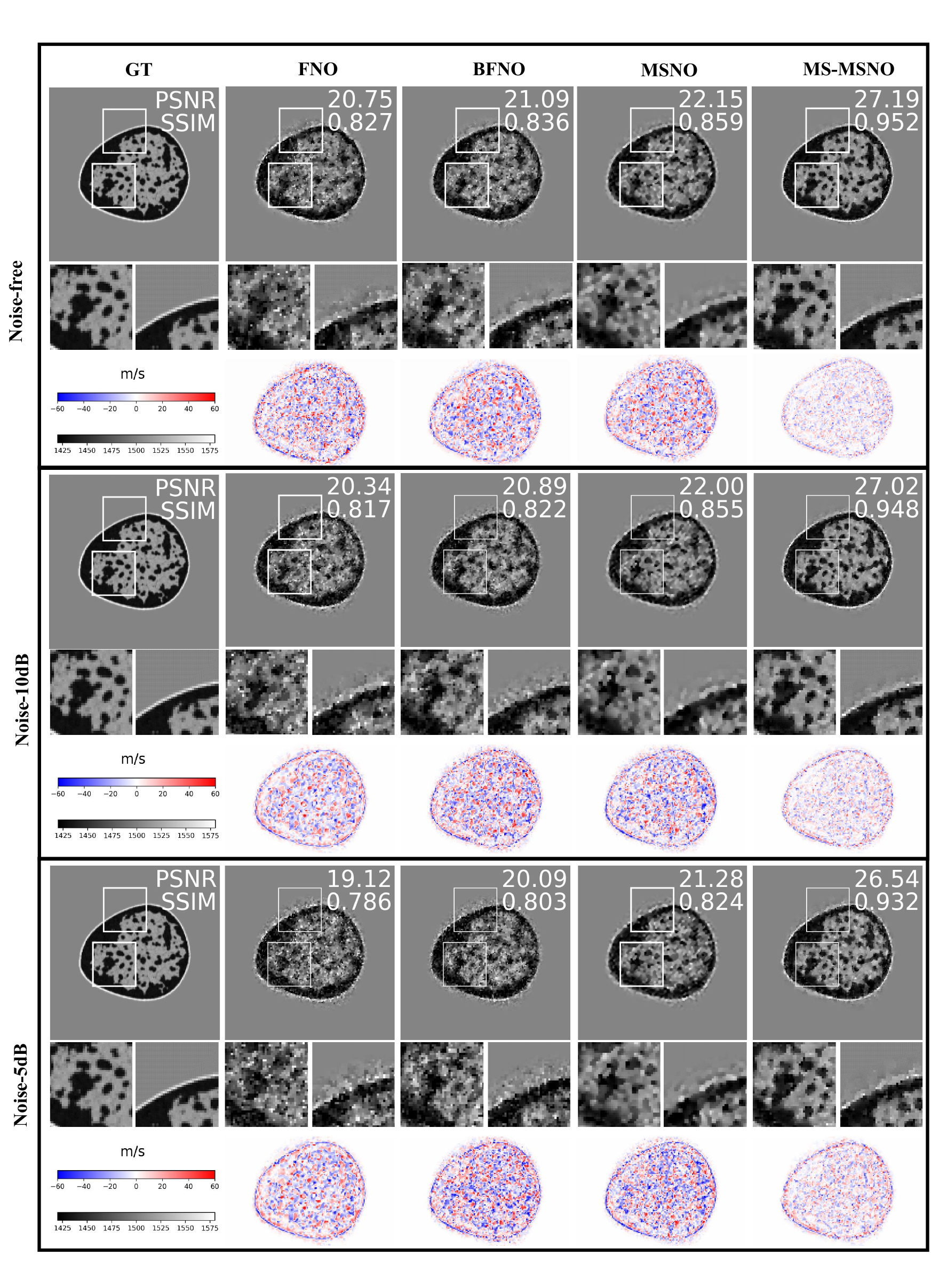} 
\caption{Qualitative reconstruction comparison on breast phantoms under noise-free, 10 dB SNR, and 5 dB SNR measurement noise. Columns show ground truth (GT), FNO, BFNO, MSNO, and our MS-MSNO. Each subplot group includes the full sound speed image, zoomed local details, and error residual map, with PSNR/SSIM metrics annotated. MS-MSNO outperforms all baselines in structural fidelity and noise robustness under all SNR levels.}    
\label{fig:case4}
\end{figure}

To assess sensitivity to measurement noise, we add Gaussian noise to the simulated measurement data and use the corrupted data as observations in the inverse problem. Figure~\ref{fig:case4} compares reconstructions under noise-free conditions and Gaussian measurement noise at signal-to-noise ratios (SNRs) of $10$ and $5\,\mathrm{dB}$ \cite{box1988signal}, with lower SNR indicating stronger noise. In the displayed examples, FNO, BFNO, and MSNO exhibit distortions in fine anatomical features and reduced visibility of glandular structures, particularly at lower SNRs. MS-MSNO shows better preservation of tissue boundaries and internal structures, together with fewer pronounced reconstruction artifacts. At $5\,\mathrm{dB}$, for example, its reconstruction retains more of the glandular patterns and tissue contours visible in the reference image than the compared single-scale models. The corresponding reconstruction-error maps also show smaller regions of large error. These qualitative results suggest that MS-MSNO is less sensitive to the tested Gaussian noise perturbations than the compared baselines.

\subsection{Impact of hyperparameters}

We investigate two key hyperparameters: the number of recursive scattering updates $N$ and the number of scale branches $s$. The former determines the depth of the latent recursion and the number of repeated wave--medium feature interactions, while the latter controls the number of scale-conditioned representations combined by the fusion module. Increasing $N$ does not correspond directly to retaining additional terms of the physical Born series, and increasing $s$ does not automatically enlarge the retained Fourier-mode set. The temperature parameter $T$ in the spectrally adaptive fusion module controls the concentration of the Softmax-based branch weights. We fix $T=5$ throughout the experiments and focus on the effects of $N$ and $s$.

To investigate the influence of $N$, we set $s=1$, corresponding to the single-scale MSNO model, and vary the number of recursive updates while keeping the remaining settings unchanged. Here, $N$ denotes the total number of latent feature updates defined in Eq.~\eqref{eq:latent_recursion}. As shown in Table~\ref{tab:ablation_study}, increasing $N$ from 3 to 7 reduces the relative $L^2$ error from $36.72\times10^{-2}$ to $10.25\times10^{-2}$. This trend indicates that additional recursive updates improve prediction accuracy over the tested range, consistent with the benefit of repeated interactions between the latent wavefield and medium features. The absolute error reduction per additional update decreases as $N$ increases, while the computational and memory costs grow. These results motivate the choice of $N=5$ as a compromise between accuracy and computational cost.

\begin{table}[ht]
\centering
\caption{Further study on computational costs and reconstruction accuracy. The top section evaluates the impact of scattering iteration number ($N$) with fixed scale numbers $s=1$, while the bottom section examines the influence of scale numbers ($s$) with fixed iteration number $N=5$.}
\label{tab:ablation_study}
\begin{tabular}{llccccc}
\toprule
\textbf{Hyper-parameters} & \textbf{Metric} & \multicolumn{5}{c}{\textbf{Configurations}} \\
\midrule
\multirow{4}{*}{\shortstack[l]{Scattering \\ Iterations}} & \(N(s=1)\)& 3 & 4 & 5 & 6 & 7 \\
\cmidrule(lr){2-7}
& Epoch(s) ($\downarrow$) & 37 & 40 & 43 & 45 & 47 \\
& Memory (GB) ($\downarrow$) & 1.2 & 3.5 & 5.8 & 7.2 & 10.6 \\
& Rel $L^2$ ($\times10^{-2}$) ($\downarrow$) & 36.72 & 20.64 & 15.33 & 12.72 & 10.25 \\
\midrule
\multirow{4}{*}{\shortstack[l]{Multi-scale \\ Factors}} & \(s(N=5)\) & 2 & 3 & 4 & 5 & 6 \\
\cmidrule(lr){2-7}
& Epoch(s) ($\downarrow$) & 40 & 50 & 62 & 75 & 90 \\
& Params (M) ($\downarrow$) & 10.2 & 14.5 & 17.7 & 25.5 & 32.3 \\
& Rel $L^2$ ($\times10^{-3}$) ($\downarrow$) & 100.12 & 62.73 & 17.15 & 6.47 & 3.51 \\
\bottomrule
\end{tabular}
\end{table}

Increasing $N$ also raises the computational and memory requirements. As $N$ increases from 3 to 7, the reported GPU memory consumption grows from $1.2\,\mathrm{GB}$ to $10.6\,\mathrm{GB}$, while the training time per epoch increases from 37 to 47 seconds. Although the update parameters are shared across iterations, deeper recursion requires additional computations and intermediate activations during training. In this architecture, $N$ specifies the number of learned latent updates rather than the number of physical scattering orders retained in a Born series. The observed error reductions suggest that repeated latent interactions improve approximation accuracy, but they do not establish how individual updates represent particular scattering contributions. Considering the diminishing absolute error reductions and increasing computational costs, $N=5$ provides a reasonable compromise under the present experimental setting.

To evaluate the effect of the number of scale branches $s$, we fix the recursion depth at $N=5$ and construct MS-MSNO models with different numbers of scale-conditioned branches. The compared models use the same recursion depth and branch architecture, while the number of branches participating in the final fusion varies. The corresponding results are reported in the bottom section of Table~\ref{tab:ablation_study}. As $s$ increases from 2 to 6, the relative $L^2$ error decreases from $100.12\times10^{-3}$ to $3.51\times10^{-3}$, with lower errors observed for each additional branch in the tested configurations. These results support the benefit of additional scale-conditioned representations, although the improvement also coincides with an increase in model capacity and cannot be attributed to scale diversity alone.

The accuracy improvement is accompanied by increased computational cost. As $s$ increases from 2 to 6, the number of trainable parameters grows from $10.2$ million to $32.3$ million, while the training time per epoch increases from 40 to 90 seconds. Thus, $s$ controls a trade-off between accuracy and computational requirements. Among the tested configurations, $s=6$ achieves the lowest relative $L^2$ error, whereas smaller values offer alternatives with fewer parameters and lower training costs.

\begin{table}[t]
\centering
\renewcommand{\arraystretch}{1.15}
\setlength{\tabcolsep}{6pt}
\caption{Comparison of training computational costs and model complexity across various frameworks. Metrics include average training time per epoch (seconds), peak GPU memory utilization (GB), and the total number of trainable parameters (Millions).}
\label{tab:comput-cost}
\begin{tabular*}{0.95\textwidth}{@{\extracolsep{\fill}} l ccc ccc @{}}
\toprule[1.2pt]
\multirow{2}{*}{\textbf{Methods}}
& \multicolumn{3}{c}{\textbf{Vanilla}}
& \multicolumn{3}{c}{\textbf{Multi-scale}} \\
\cmidrule(lr){2-4} \cmidrule(lr){5-7}
& \textbf{FNO} & \textbf{BFNO} & \textbf{MSNO}
& \textbf{FNO} & \textbf{BFNO} & \textbf{MSNO} \\
\midrule[1.0pt]
\textbf{Epoch(s)} ($\downarrow$)
& 62 & 45 & 43 & 78 & 63 & 60 \\
\addlinespace[2pt]
\textbf{Memo.(GB)} ($\downarrow$)
& 12.2 & 6.3 & 5.8 & 23.5 & 12.3 & 11.6 \\
\addlinespace[2pt]
\textbf{Params(M)} ($\downarrow$)
& 18.7 & 6.2 & 5.7 & 52.8 & 18.9 & 17.7 \\
\bottomrule[1.2pt]
\end{tabular*}
\end{table}

\subsection{Comparison of computational efficiency} 

Table~\ref{tab:comput-cost} reports the average training time per epoch, GPU memory consumption, and number of trainable parameters for FNO, BFNO, MSNO, and their multiscale variants in the breast-phantom experiment. In the FNO implementation considered here, each Fourier layer has its own parameters. In contrast, BFNO and MSNO share the update parameters across recursive iterations, reducing the parameter count. Both models also exhibit lower GPU memory consumption than FNO in the reported experiments, although memory requirements depend on intermediate activations and implementation details as well as parameter count.

Introducing multiple scale-conditioned branches increases the parameter count relative to the corresponding single-scale models. However, MS-BFNO and MS-MSNO retain parameter sharing across recursive updates within each branch. Despite the additional branches, MS-MSNO has a parameter count and training time per epoch comparable to those of the single-scale FNO baseline. Specifically, MS-MSNO contains $17.7$ million trainable parameters, compared with $18.7$ million for FNO. Its average training time is 60 seconds per epoch, compared with 62 seconds for FNO under the same experimental setting.

At these comparable parameter counts and training costs, MS-MSNO achieves a lower relative $L^2$ error and higher PSNR and SSIM values in the reported accuracy comparisons. The visual results also show improved delineation of breast-tissue boundaries and fine-scale structures. Taken together, these results indicate a favorable accuracy--cost trade-off for MS-MSNO in the tested configuration. They support combining multiscale input conditioning with shared recursive updates, while not implying that parameter sharing alone accounts for the observed accuracy or runtime differences.

\begin{figure}[h]    
\centering    
\includegraphics[width=1\linewidth,height=0.5\textheight,keepaspectratio]{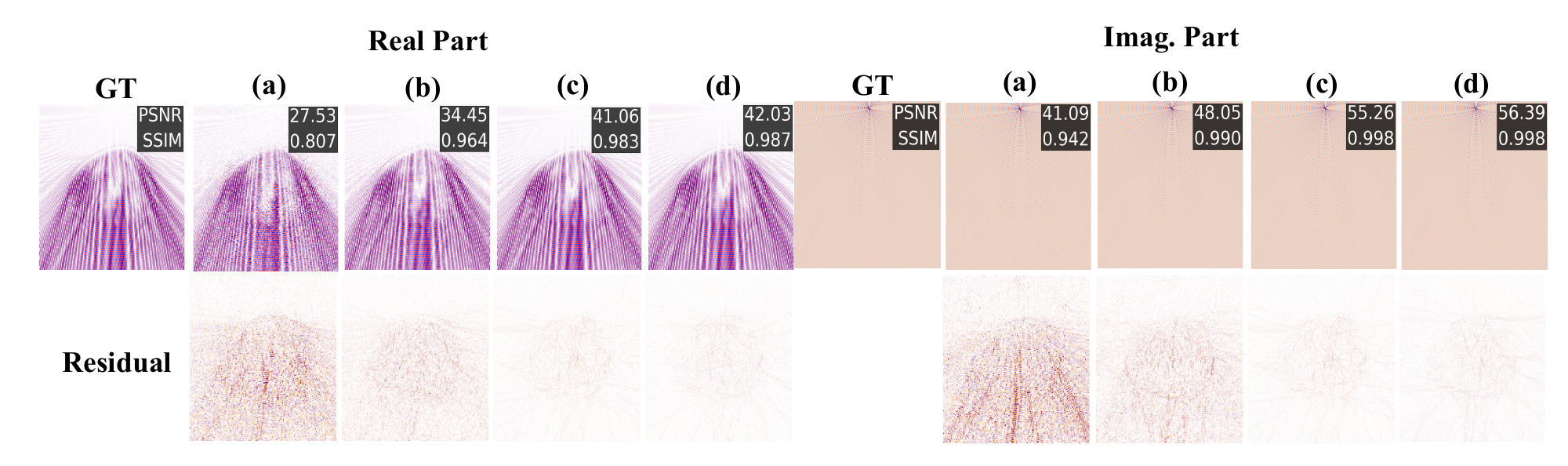}    
\caption{
Ablation study on the forward wavefield prediction task. 
The four compared configurations are: 
(a) single-step MSNO with \(N=1, s=1\); 
(b) recursive MSNO with \(N=5, s=1\); 
(c) MS-MSNO with equal-weight branch fusion, \(N=5, s=5, \gamma_{b}=\frac{1}{5}\); 
and (d) MS-MSNO with spectrally adaptive fusion, \(N=5, s=5\). 
The first column shows the ground-truth wavefields, while the remaining columns show the predicted real and imaginary parts together with their corresponding residual maps. }
\label{fig:Ablation_forward}
\end{figure}

\begin{figure}[h]    
\centering    
\includegraphics[width=0.9\linewidth]{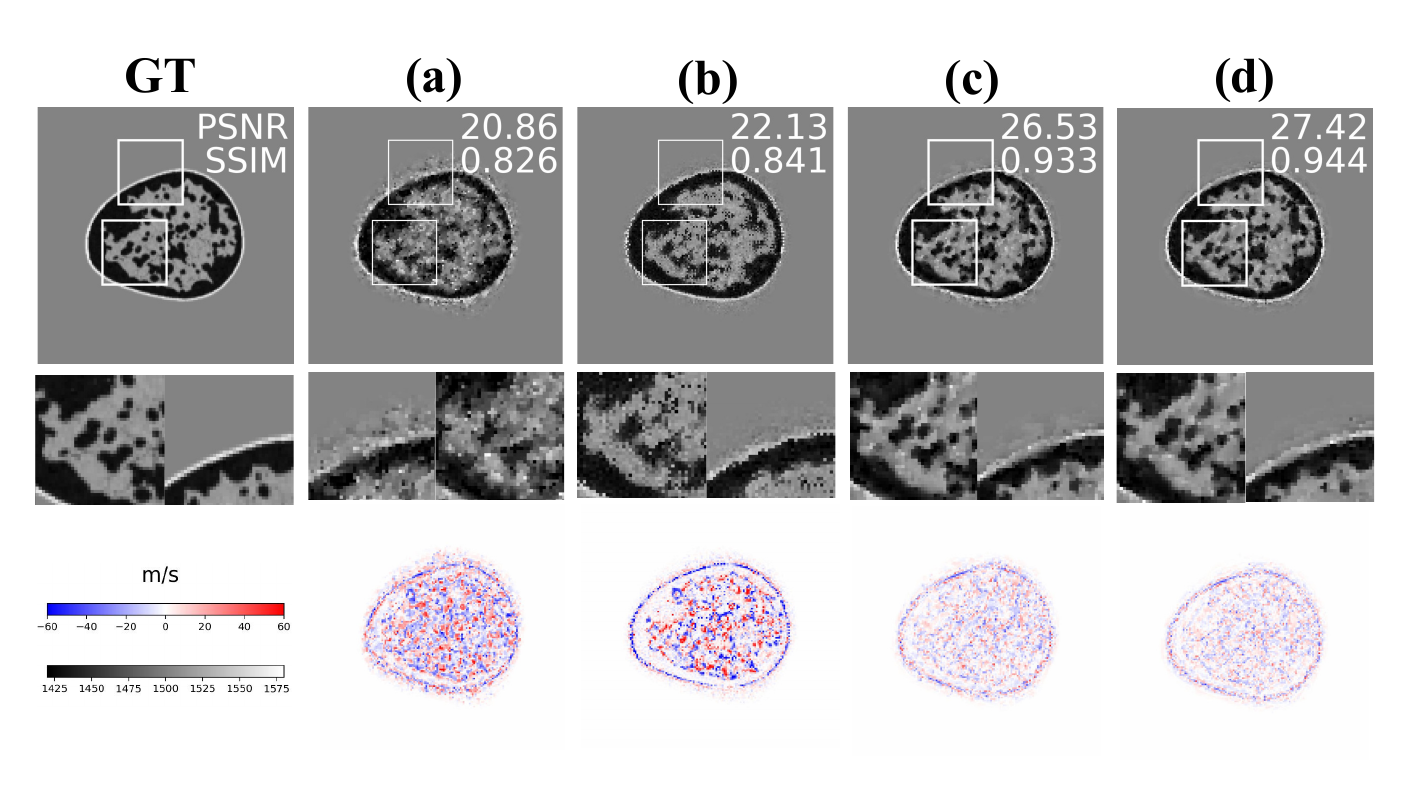}   
\caption{
Ablation study on the inverse medium reconstruction task. 
The four compared configurations are: 
(a) single-step MSNO with \(N=1, s=1\); 
(b) recursive MSNO with \(N=5, s=1\); 
(c) MS-MSNO with equal-weight branch fusion, \(N=5, s=5, \gamma_{b}=\frac{1}{5}\); 
and (d) MS-MSNO with spectrally adaptive fusion, \(N=5, s=5\). 
The first column presents the ground-truth sound-speed map, and the remaining columns show the reconstructed results under different configurations. 
The zoomed-in regions highlight local structural details, and the bottom row presents the corresponding reconstruction residuals.}  
\label{fig:Ablation_inverse}
\end{figure}

\subsection{Ablation studies}

We conduct ablation studies on forward wavefield prediction and inverse medium reconstruction to examine the effects of the number of latent recursive updates $N$, the number of scale branches $s$, and the spectrally adaptive fusion (SAF) module. The results are presented in Figures~\ref{fig:Ablation_forward} and~\ref{fig:Ablation_inverse}. Four configurations are compared: (a) single-step MSNO with $N=1$ and $s=1$; (b) recursive MSNO with $N=5$ and $s=1$; (c) MS-MSNO with equal-weight fusion, using $N=5$, $s=5$, and $\gamma_b=1/s$; and (d) MS-MSNO with SAF, using $N=5$ and $s=5$.

The comparison between (a) and (b) assesses the effect of additional latent recursive updates. Increasing $N$ improves wavefield prediction and reduces reconstruction artifacts in the displayed examples. Because the update parameters are shared, this comparison increases the recursion depth without introducing additional update parameters, although the computational cost grows. The comparison between (b) and (c) assesses the combined effect of introducing multiple scale-conditioned branches and increasing model capacity. Configuration (c) produces sharper structural reconstructions and better preserves fine-scale details, but this comparison alone does not separate the benefit of scale conditioning from that of the additional parameters. Finally, the comparison between (c) and (d) evaluates SAF against equal-weight averaging with the same branch count and recursion depth. The sample-dependent, spectral-response-based weights yield further improvements in the reported metrics and reduce visible local reconstruction errors. Overall, the improvements in PSNR and SSIM, together with the prediction- and reconstruction-error maps, support the contributions of recursive updates, multiscale branches, and SAF to the performance of the full model in the tested settings.

\section{Conclusion}\label{sec6}

In this work, we proposed a latent-space Multiple Scattering Neural Operator (MSNO) for forward wavefield prediction and inverse medium reconstruction. Inspired by the recursive structure of the Born iteration for the Helmholtz equation, MSNO combines spectral interactions with shared latent updates to model repeated wave--medium interactions. The architecture follows this scattering structure without identifying its learned updates with individual terms of the physical Born series. An auxiliary autoencoder loss encourages encoder--decoder reconstruction consistency for the incident field, complementing the total-field prediction objective. We further developed a multiscale extension, MS-MSNO, that combines scale-conditioned branches with a spectrally adaptive fusion (SAF) module. SAF computes nonnegative, normalized branch weights from the radial spectral responses of the current branch predictions. These weights are recomputed for each input during training and inference, including at each inversion iteration as the medium estimate changes. This mechanism provides sample-dependent aggregation of the branch outputs and supports the recovery of fine-scale wavefield and medium structures. Experiments on the tested electromagnetic and acoustic datasets show that MSNO and MS-MSNO improve forward prediction and inverse reconstruction relative to the compared FNO and BFNO baselines. The spectral analyses and ablation studies support the benefits of recursive latent updates, multiscale representations, and adaptive fusion in these settings. MS-MSNO also achieves a favorable balance between accuracy and computational cost, with parameter counts and training times comparable to the single-scale FNO baseline in the reported breast-phantom experiment. These empirical results do not establish convergence of the learned recursion or generalization beyond the tested physical and acquisition settings.

Future work will investigate lighter architectures, generalization across frequencies and medium distributions, and extensions to other wave-propagation models. Hybrid schemes that combine learned surrogates with conventional numerical solvers also offer a direction for improving physical consistency and controlling approximation errors during inversion.

\section{Acknowledgements}
The work is supported by the National Natural Science Foundation of China (NSFC 12671658 and T2541053).

\bibliographystyle{unsrt}
\bibliography{references}

\end{document}